\documentclass{article}
\usepackage[T1]{fontenc}
\usepackage{bimuon_arxiv,times}
\usepackage{amsmath,amssymb,amsthm,mathtools,enumitem,booktabs,graphicx,float,placeins,needspace}
\newcounter{algorithm}
\newcounter{algline}
\newcounter{algdepth}
\makeatletter
\def\fps@algorithm{tbp}
\def\ftype@algorithm{4}
\def\ext@algorithm{loa}
\newenvironment{algorithm}[1][tbp]{\@float{algorithm}[#1]\hrule\smallskip}{\par\smallskip\hrule\end@float}
\makeatother
\newcommand{\algcaption}[1]{\refstepcounter{algorithm}\noindent\textbf{Algorithm \thealgorithm}\quad #1\par\smallskip\hrule\smallskip}
\newenvironment{algorithmic}[1][]{\setcounter{algline}{0}\setcounter{algdepth}{0}\begin{list}{}{\setlength{\leftmargin}{1.8em}\setlength{\labelwidth}{1.4em}\setlength{\labelsep}{.4em}\setlength{\itemsep}{2pt}\setlength{\parsep}{0pt}\setlength{\topsep}{2pt}}}{\end{list}}
\newcommand{\State}{\stepcounter{algline}\item[\small\thealgline:]\hspace*{\dimexpr\value{algdepth}em\relax}}
\newcommand{\Require}{\item[]\textbf{Input:} }

\newcommand{\For}[1]{\State\textbf{for} #1 \textbf{do}\stepcounter{algdepth}}
\newcommand{\EndFor}{\addtocounter{algdepth}{-1}\State\textbf{end for}}
\newcommand{\Return}{\textbf{return} }

\usepackage{hyperref,url}
\usepackage{xcolor,xspace,wrapfig}
\usepackage{tcolorbox}

\allowdisplaybreaks[2]
\newtheorem{assumption}{Assumption}
\newtheorem{theorem}{Theorem}
\newtheorem{proposition}{Proposition}
\newtheorem{lemma}{Lemma}
\newtheorem{corollary}{Corollary}
\theoremstyle{remark}
\newtheorem{remark}{Remark}
\theoremstyle{plain}
\newcommand{\E}{\mathbb E}
\newcommand{\F}{\mathrm F}
\newcommand{\op}{\mathrm{op}}
\newcommand{\NS}{\operatorname{NS}}
\newcommand{\polar}{\operatorname{polar}}

\newcommand{\alg}{$\mathsf{BiMuon}$\xspace}
\newcommand{\algn}{$\mathsf{BiMuon}$}
\title{\alg: A Geometry-Aware Stochastic Bilevel Optimization Approach}
\author{Zhen Qin \quad Zhiyao Zhang \quad Yingbin Liang \quad Jia Liu\\
\normalfont Department of Electrical and Computer Engineering\\
\normalfont The Ohio State University}
\date{}
\hypersetup{
  pdftitle={BiMuon: A Geometry-Aware Stochastic Bilevel Optimization Approach},
  pdfauthor={Zhen Qin, Zhiyao Zhang, Yingbin Liang, Jia Liu},
  hidelinks
}

\DeclareMathOperator*{\argmin}{arg\,min}
\newtheorem{example}{Example}

\begin{document}
\maketitle
\begin{abstract}
Stochastic bilevel optimization provides a general framework for many machine learning applications, such as hyperparameter learning, data reweighting, language model alignment, and actor-critic reinforcement learning.
However, most existing bilevel optimization methods treat matrix-valued gradients as flattened vectors and perform optimization in the resulting vectorized space, ignoring the matrix structures common in modern neural networks.
Moreover, the lower-level (LL) problem curvature can induce significant spectral imbalance in the upper-level (UL) geometry.
Such limitations of existing bilevel optimization methods motivate us to develop \alg, a geometry-aware stochastic bilevel optimization method that integrates implicit hypergradient estimation in bilevel optimization with Muon-style orthogonalized updates at the UL.
To our knowledge, \alg is the first stochastic BLO method with Muon-style orthogonalized matrix updates and an $O(\epsilon^{-4})$ iteration-complexity guarantee, matching the best-known rates achieved by Muon-type methods based on matrix parameterizations.
To improve computational and parallel efficiency for models with multiple matrix parameters, we further study \alg for blockwise diagonal matrix approximation and derive a relative stepsize allocation rule based on UL progress and blockwise LL sensitivity.
We evaluate \alg on representation learning and blockwise bilevel optimization, providing empirical support for our methods.\footnote{Code: \url{https://github.com/ZhenQinZoe/BiMuon}.}
\end{abstract}
\section{Introduction}
\label{sec:intro}

{\bf 1) Background and Motivation}
Stochastic bilevel optimization (BLO) is a unified foundational optimization framework for many machine learning applications.
Representative examples include reinforcement learning from human feedback (RLHF)~\citep{chakraborty2023parl}, data reweighting for large-scale pretraining~\citep{kang2025autoscale}, preference-based fine-tuning~\citep{ding2024sail}, and hyperparameter optimization~\citep{franceschi2018bilevel,lorraine2020implicit}.
These problems naturally admit a BLO formulation, where an upper-level (UL) problem governs a set of control variables, while a lower-level (LL) problem defines the training dynamics of the base model.

In recent years, significant progress has been made in BLO, with algorithms based on implicit differentiation, Neumann-series approximations, and stochastic optimization achieving provable convergence guarantees under nonconvex--strongly-convex assumptions \citep{franceschi2018bilevel,lorraine2020implicit,ji2021bilevel,chen2022stable,dagreou2022framework}.
However, almost all BLO methods adopt a flattened-vector-based (FVB) approach to optimize model parameters, vectorizing weight matrices and applying coordinate-wise updates.
While the FVB approach is mathematically convenient, it ignores the matrix structures that are ubiquitous in neural network models.

Meanwhile, in contemporary neural architecture training with BLO (e.g., large language models and many Transformer-based architectures), the UL and LL variables can naturally be matrix-based linear operators whose training dynamics exhibit strong spectral anisotropy and low-rank structure \citep{liu2025scalable,zhao2024galore}.
Empirical and theoretical studies consistently show that gradients and update matrices in such models can be highly {\em ill-conditioned,} with learning dynamics governed by a small number of dominant singular directions~\citep{jordan2024muon,zhao2024galore}.
As a result, FVB updates tend to over-emphasize directions associated with large singular values, while making limited progress along flatter or smaller spectral directions.
In BLO, this mismatch can be further exacerbated by the LL curvature: as we show theoretically in this paper, poor LL conditioning can slow LL optimization and, through the bilevel composition, induce stronger spectral imbalance in the UL geometry and hypergradient.

Interestingly, recent advances in matrix-based optimization based on singular-spectrum regularization, most notably Muon \citep{jordan2024muon} and its variants, demonstrate that orthogonalized updates can better {\em align optimization dynamics with the spectral structure of neural network matrix parameters} and achieve strong empirical performance \citep{liu2025scalable}.
Specifically, by normalizing singular values, Muon mitigates the dominance of large spectral components and promotes balanced updates across subspaces.
Moreover, recent theoretical results have begun to establish convergence guarantees for Muon-style orthogonalized updates in single-level stochastic optimization~\citep{shen2025convergence,kim2026muon}.
These recent advances in singular-spectrum-based optimizers naturally raise the following fundamental question:

\begin{tcolorbox}[left=1.2pt,right=1.2pt,top=1.2pt,bottom=1.2pt,before skip=6pt,after skip=6pt]
\textbf{(Q): Can we develop a Muon-style matrix-based algorithm for geometry-aware stochastic bilevel optimization with provable convergence guarantees?}
\end{tcolorbox}

{\bf 2) Technical Challenges and Major Contributions:}
Unfortunately, developing singular-spectrum-regularized approaches for BLO is highly non-trivial due to several technical challenges unseen in single-level settings.
In BLO, hypergradient computation involves LL Hessian inversion, and LL and UL variables are coupled across iterations.
Specifically, each UL update changes the UL variable, thus altering the LL solution that must be tracked by a finite LL optimization process.
The resulting LL tracking error in turn affects the subsequent hypergradient estimate, creating feedback loops between matrix-based UL and LL updates~\citep{ji2021bilevel,grazzi2021stochastic}.
Moreover, practical Muon implementations often use a finite Newton--Schulz (NS) approximation instead of the exact polar decomposition, incurring an additional layer of approximation errors~\citep{kim2026muon}.

In this work, we address the aforementioned challenges by proposing \emph{\alg}, which, to our knowledge, is the first geometry-aware, matrix-based stochastic BLO algorithm built on Muon-style orthogonalized updates.
Our key idea is to decouple bilevel hypergradient estimation from orthogonalized UL updates: the LL problem is first tracked using warm-started stochastic gradient descent (SGD), while the UL hypergradient is estimated using a truncated Neumann-series approximation; orthogonalized updates are then applied at the final UL parameter update to exploit matrix geometry.
This design allows \alg to retain the standard stochastic bilevel estimation procedure, while enabling matrix-based optimization principles within bilevel learning.
Our contributions are threefold:

\begin{itemize}[leftmargin=*]
\item We first theoretically show that LL curvature in BLO can induce significant {\em spectral imbalance} in the UL geometry, highlighting the compelling need for geometry-aware optimization in BLO. To mitigate this spectral imbalance, we propose \textbf{\alg}, the first geometry-aware stochastic BLO method with Muon-style orthogonalized matrix updates.

\item We then establish the finite-time convergence rate guarantees for both the exact-polar and finite-NS versions of \alg.
Specifically, both exact-polar and finite-NS versions of \alg achieve a finite-time convergence rate of $\mathbb{E}\|\nabla\Phi(X_R)\|_*=O(T^{-1/4})$,
where $T$ is the maximum number of iterations.
For any given error tolerance $\epsilon>0$, this convergence rate implies an $\widetilde{O}(\epsilon^{-4})$ sample complexity, matching the best-known results for stochastic single-level Muon-type optimization.

\item To improve computational and parallel efficiency for models with multiple matrix parameters, we further study \alg with blockwise diagonal matrix approximation and propose a step-size allocation rule based on UL progress and blockwise LL sensitivity.
We further evaluate \alg on representation learning and blockwise bilevel optimization problems, which empirically verify the performance of our proposed \alg algorithmic designs.
\end{itemize}

\section{Related Work}

In this section, we provide an overview of two lines of research closely related to our work.

\textbf{1) Stochastic Bilevel Optimization.}
Gradient-based BLO methods commonly use implicit or iterative differentiation to approximate hypergradients under the strongly convex LL problem assumption~\citep{franceschi2018bilevel,ghadimi2018approximation,ji2021bilevel,yang2021faster}.
Recent stochastic methods further improve LL tracking and hypergradient estimation through single-timescale and variance-reduced schemes \citep{chen2022stable,dagreou2022framework,chu2024spaba}.
These works primarily study how to estimate and track the hypergradient, while the resulting UL update typically remains flattened-vector-based (FVB).
Bregman and Riemannian BLO methods introduce alternative optimization geometries \citep{huang2022bregman,han2024framework,li2025riemannian}.
In contrast, our \alg method targets matrix-valued UL variables and exploits the singular-spectrum structure of their hypergradients through Muon-style orthogonalized updates.

\textbf{2) Muon and Other Matrix-Aware Optimizers.}
Matrix-aware optimizers exploit structure beyond flattened parameter vectors.
For example, Shampoo and SOAP use matrix preconditioning \citep{gupta2018shampoo,vyas2024soap}, while GaLore exploits low-rank gradient structure \citep{zhao2024galore}.
Most notably, Muon orthogonalizes gradient or momentum matrices by normalizing their singular spectrum \citep{jordan2024muon}.
Recent works on Muon have verified large-scale empirical performance \citep{liu2025scalable} and established theoretical convergence guarantees for exact and finite-NS updates \citep{shen2025convergence,kim2026muon}.
Norm-based steepest descent and Scion further connect matrix updates to norm geometry \citep{bernstein2024old,pethick2025training}.
However, all of these results focus on single-level optimization and do not address LL tracking or implicit hypergradient errors.
Blockwise matrix methods motivate processing large parameter matrices separately \citep{amari2019fisher,mendlerdunner2020randomized,bahamou2023mini}; our blockwise design additionally uses LL sensitivity to allocate relative stepsizes across blocks.

\section{Problem Statement and Preliminaries}
\label{sec:preliminaries}
In this section, we first review the standard FVB formulation of BLO, which is followed by an in-depth analysis of the significant bilevel-inherent spectral imbalance to motivate our subsequent \alg algorithmic design.
For matrices in this paper, we let
$\langle A,B\rangle=\operatorname{tr}(A^\top B)$,
$\|A\|_F=\sqrt{\langle A,A\rangle}$, and let
$\|A\|_*$ and $\|A\|_{\mathrm{op}}$ denote the nuclear and spectral norms.
For matrix pairs,
$\|(X,Y)\|_F^2:=\|X\|_F^2+\|Y\|_F^2$.
For an $m\times n$ UL matrix, $r=\min\{m,n\}$ is the rank cap.

\subsection{BLO Problem Statement and the FVB Hypergradient-Based Approaches}
\label{sec:problem}
\textbf{1) Problem Statement.}
In the FVB form, a stochastic BLO problem can generally be written as
\begin{equation*}
\begin{aligned}
\min_{x\in\mathbb R^{d_x}}\quad &\Phi(x):=f(x,y^\star(x))=\mathbb E_\xi[F(x,y^\star(x);\xi)]\\
\text{s.t.}\quad &y^\star(x) \in \argmin_{y\in\mathbb R^{d_y}}g(x,y):=\mathbb E_\zeta[G(x,y;\zeta)].
\end{aligned}
\end{equation*}
Here, $x$ denotes the UL variable and $y^\star(x)$ is an optimizer of the LL objective $g(x,\cdot)$ with $x$ being fixed.
We follow the standard BLO setting in the literature, where the UL objective $\Phi$ is non-convex in $x$ and the LL objective $g(x,\cdot)$ is $\mu$-strongly convex for some $\mu>0$, ensuring that the LL solution $y^\star(x)$ is uniquely defined.
The expectations in the UL and LL objectives are taken over the distributions of random variables $\xi$ and $\zeta$, respectively.

\textbf{2) The FVB-Based Hypergradient Approach for BLO.}
When the LL problem is strongly convex, under the differentiability assumptions (cf. Section~\ref{sec:theory}), the UL objective $\Phi(x)=f(x,y^\star(x))$ admits a well-defined hypergradient.
Specifically, by the implicit function theorem~\citep{ghadimi2018approximation}, the hypergradient of $\Phi$ can be written as:
\begingroup
\begin{equation}\label{eq:exact_hypergradient}
\nabla\Phi(x)=\nabla_x f(x,y^\star(x))-\nabla_{xy}^2 g(x,y^\star(x))\bigl[\nabla_{yy}^2g(x,y^\star(x))\bigr]^{-1}\nabla_y f(x,y^\star(x)).
\end{equation}
\endgroup
However, direct computation of the inverse-Hessian action in~\eqref{eq:exact_hypergradient} is generally expensive.
A standard approach in stochastic BLO is to use the truncated
Neumann-series approximation~\citep{ji2021bilevel,yang2021faster}
$
H^{-1}
\approx
\eta\sum_{i=0}^{Q}(I-\eta H)^i,
$
where $H=\nabla_{yy}^2g(x,y)$ is the LL Hessian at the current point and $\eta>0$ is chosen such that
$\|I-\eta H\|_{\op}<1$.
This approximation can be evaluated using Hessian--vector products without forming the Hessian inverse explicitly.
We use this standard construction for the hypergradient estimator in Section~\ref{sec:bimuon-method}.

The FVB formulation can directly extend to matrix-valued variables
$X\in\mathbb R^{m\times n}$ and $Y\in\mathbb R^{h\times s}$ via
$x=\operatorname{vec}(X)$ and $y=\operatorname{vec}(Y)$, with
$d_x=mn$ and $d_y=hs$, provided the updates remain entry-wise.
In matrix form, $\nabla_{XY}^2 g$ is the adjoint of
$D_X(\nabla_Y g)$ and maps LL matrices to UL matrices; however, such a matrix-valued extension does {\em not} leverage matrix structures (e.g., spectral properties or low-rank geometry) that arise in modern neural networks.
We next use a quadratic bilevel example to show how LL curvature can inherently induce {\em significant spectral imbalance} in the UL geometry.

\subsection{Bilevel-Inherent Spectral Imbalance: A Quadratic Case Study}
\label{sec:bilevel-spectral-imbalance}

Motivated by the fact that many nonlinear functions of practical interest can be locally well approximated by quadratic functions, we use a quadratic bilevel problem to illustrate how differences in LL curvature can be {\em amplified} through bilevel composition, resulting in highly unequal singular values in the UL hypergradient.
Let $(Z_{\mathrm{tr}},R_{\mathrm{tr}})$ and $(Z_{\mathrm{val}},R_{\mathrm{val}})$ denote the training and validation data with $n_{\mathrm{tr}}$ and $n_{\mathrm{val}}$ samples, respectively.
For matrix variables $X,Y\in\mathbb{R}^{m\times n}$, consider the following quadratic UL and LL functions:
\begingroup
\begin{equation}\label{eq:quadratic-motivation}
f(X,Y)=\tfrac{\|Z_{\mathrm{val}}Y-R_{\mathrm{val}}\|_\F^2}{2n_{\mathrm{val}}}+\tfrac{\gamma}{2}\|X-X_{\mathrm{prior}}\|_\F^2,
\quad
g(X,Y)=\tfrac{\|Z_{\mathrm{tr}}Y-R_{\mathrm{tr}}\|_\F^2}{2n_{\mathrm{tr}}}+\tfrac{\mu}{2}\|Y-X\|_\F^2,
\end{equation}
\endgroup
where $\mu>0$ controls the coupling between the UL and LL variables, $\gamma\geq0$ is the UL regularization parameter, and $X_{\mathrm{prior}}$ is a reference matrix.
The LL problem adapts $Y$ using the training data while keeping it close to $X$, whereas the UL objective evaluates the adapted model on validation data.

\begin{wrapfigure}{r}{.32\textwidth}
\vspace{-.1in}
\centering
\includegraphics[width=\linewidth,trim=0 6pt 0 2pt,clip]{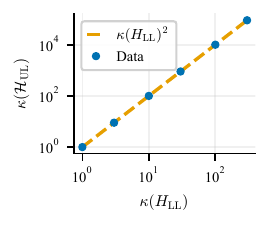}
\vspace{-.3in}
\caption{$\kappa(\mathcal H_{\mathrm{UL}})=\kappa(H_{\mathrm{LL}})^2$.}
\label{fig:spectral-illustration}
\vspace{.1in}
\end{wrapfigure}

Define the Hessian matrix that characterizes LL curvature as $H_{\mathrm{LL}} :=Z_{\mathrm{tr}}^\top Z_{\mathrm{tr}}/n_{\mathrm{tr}}+\mu I$.
Under isotropic validation curvature $Z_{\mathrm{val}}^\top Z_{\mathrm{val}}/n_{\mathrm{val}}=I$ and $\gamma=0$, it can be readily verified that the induced UL Hessian operator $\mathcal{H}_{\mathrm{UL}}:=\nabla^2\Phi(X)$ satisfies:
\[
\mathcal{H}_{\mathrm{UL}}[\Delta X]
=
\mu^2H_{\mathrm{LL}}^{-2}\Delta X,
\qquad
\kappa(\mathcal{H}_{\mathrm{UL}})
=
\kappa(H_{\mathrm{LL}})^2,
\]
where $\kappa(\cdot)$ is the condition number (i.e., the ratio of the largest to the smallest eigenvalue).
Thus, curvature differences across LL directions are amplified {\bf quadratically} through the bilevel composition.
To visualize this condition number amplification, we construct controlled instances of~\eqref{eq:quadratic-motivation} with prescribed LL Hessian spectra, which isolate the analytical spectral effect with full details in Appendix~\ref{app:controlled-spectral}.
Figure~\ref{fig:spectral-illustration} illustrates the squared-conditioning effect as the LL Hessian becomes more spectrally imbalanced.

More importantly, this amplification can appear directly in the UL hypergradient.
Let $h_i$ denote an eigenvalue of $H_{\mathrm{LL}}$ and $x_i$ the corresponding component of a UL displacement aligned with the LL curvature eigendirections.
The associated hypergradient singular magnitude scales as $x_i/h_i^2$.
Appendix~\ref{app:controlled-spectral} further illustrates this amplification and the effect of polar normalization in Figure~\ref{fig:spectral-supplement}.

Hence, anisotropic LL curvature can produce a UL hypergradient with highly unequal singular values, causing a FVB UL update to be dominated by a small number of spectral directions.
Complete derivations are provided in Appendix~\ref{app:quadratic}.
As mentioned above, quadratic BLO provides a local second-order view of more general smooth bilevel problems, where the UL Hessian characterizes the geometry up to higher-order terms.
This motivates our subsequent Muon-style orthogonalized UL updates in the next section, which regulate the singular spectrum to mitigate the spectral imbalance.

\section{The Proposed \alg Algorithm}
\label{sec:algorithm}

In this section, we first introduce the necessary background of the Muon optimizer in Section~\ref{sec:muon}, which facilitates subsequent discussions of our \alg design in Section~\ref{sec:bimuon-method}.

\subsection{The Muon Optimizer: A Primer}
\label{sec:muon}
Consider the single-level optimization
$\min_{X\in\mathbb{R}^{m\times n}}\phi(X)$.
Given a nonzero matrix $M\in\mathbb{R}^{m\times n}$ with compact singular
value decomposition (SVD)
$
M = U\Sigma V^\top,
$
Muon ~\citep{jordan2024muon} constructs an orthogonalized update direction through the supported
polar factor
$
\operatorname{polar}(M):=UV^\top,$ where $
\operatorname{polar}(0):=0.
$
A Muon-style update takes the form
\[
X_{t+1}
=
X_t-\beta_t\operatorname{polar}(M_t),
\]
where $M_t$ is the gradient or momentum matrix.
The polar transformation preserves the active left and right singular subspaces of $M_t$, while removing its positive singular-value scaling.

However, computing the exact polar factor requires expensive SVD computations.
In practical Muon implementations, the polar factor is approximated by Newton--Schulz (NS) iterations based only on matrix multiplications.
Following \citet{kim2026muon}, we initialize
$
O^{[0]}
=
\frac{M}{\max\{1,\|M\|_\F\}},
$
and apply
\[
O^{[j+1]}
=
p\!\left(O^{[j]}(O^{[j]})^\top\right)O^{[j]},
\qquad
j=0,\ldots,q-1,
\]
where $p$ is a degree-$d$ Taylor approximation of $x^{-1/2}$ and $q$ is the NS iteration depth.
We denote the output after $q$ iterations by
$
\operatorname{NS}(M,q):=O^{[q]}.
$
The explicit polynomial and the complete definition of $\NS(M,q)$ are given in Appendix~\ref{app:algorithm-details}, with the routine summarized in Algorithm~\ref{alg:ns}.
We note that the exact polar factor provides a useful reference transformation for theoretical understanding, while the finite-NS technique gives a practical approximation using only matrix multiplications.
We next incorporate these matrix-based directions into stochastic bilevel optimization.

\subsection{The Proposed \alg Algorithm}
\label{sec:bimuon-method}
Our \alg method combines the following steps: warm-started LL tracking, stochastic hypergradient
estimation, and a Muon-style UL update, which are also summarized in Algorithm~\ref{alg:bimuon}.

\begin{algorithm}[t]
\algcaption{\alg: Exact-Polar and Newton--Schulz Variants}
\label{alg:bimuon}
\begin{algorithmic}[1]
\Require Horizon $T$; depths $D,Q,q$; stepsizes $\alpha,\beta,\eta$; momentum $\theta$;
initial points $X_0,Y_0$; batch sizes $B_{\mathrm{low}},B_f,B_{xy},B_H$
\State Initialize $M_{-1}=0$
\For{$k=0,\ldots,T-1$}
  \State Set $Y_k^0=Y_0$ if $k=0$, and $Y_k^0=Y_{k-1}^D$ otherwise
  \For{$t=0,\ldots,D-1$}
    \State Draw a fresh batch $\mathcal S_{k,t}$ of size $B_{\mathrm{low}}$; update $Y_k^{t+1}$ using~\eqref{eq:ll_update}
  \EndFor
  \State Draw independent fresh batches $\mathcal D_{F,k},\mathcal D_{G,k},\{\mathcal B_{k,j}\}_{j=1}^Q$ as specified below
  \State Estimate $\widehat{\nabla\Phi}(X_k)$ using~\eqref{eq:stochastic_hg} via the Neumann recursion
  \State Update $M_k$ using~\eqref{eq:momentum_update}
  \State \textbf{Option I} (Exact-Polar): $O_k=\operatorname{polar}(M_k)$; \quad
    \textbf{Option II} (Finite-NS): $O_k=\operatorname{NS}(M_k,q)$
  \State $X_{k+1}=X_k-\beta O_k$
\EndFor
\State \textbf{return} $X_T$
\end{algorithmic}
\end{algorithm}
\textbf{1) LL Trajectory Tracking.}
At the LL, our goal is to track the solution $Y^\star(X_k)$ of the LL problem.
At outer iteration $k\ge1$, the LL optimization is initialized using the previous LL output, $Y_k^0 = Y_{k-1}^D$, with $Y_0^0=Y_0$.
Starting from $Y_k^0$, we perform $D$ stochastic gradient steps:
\begin{equation}\label{eq:ll_update}
    Y_k^{t+1}
    =
    Y_k^t
    -
    \alpha
    \nabla_Y G(X_k,Y_k^t;\mathcal{S}_{k,t}),
    \qquad
    t=0,\ldots,D-1,
\end{equation}
where $\mathcal{S}_{k,t}$ is a fresh mini-batch of size $B_{\mathrm{low}}$.
The resulting $Y_k^D$ serves as the approximate LL solution used in the UL hypergradient estimation.

The warm start is important in the bilevel setting because the target $Y^\star(X_k)$ changes as the UL variable is updated.
Thus, instead of solving each LL problem from scratch, \alg initializes the next LL loop from the previous LL output and tracks the moving solution across outer iterations.

\textbf{2) Stochastic UL Hypergradient Updates.}
At the UL, we update $X$ using a stochastic hypergradient estimator based on a truncated Neumann-series approximation.
At $(X_k,Y_k^D)$, independent fresh mini-batches $\mathcal D_{F,k}$, $\mathcal D_{G,k}$, and $\{\mathcal B_{k,j}\}_{j=1}^{Q}$, of sizes $B_f$, $B_{xy}$, and $B_H$, respectively, are used to compute the partial gradients and Hessian--vector products.
This yields:
\begin{equation}\label{eq:stochastic_hg}
\begin{aligned}
\widehat{\nabla\Phi}(X_k)
={}&\nabla_XF(X_k,Y_k^D;\mathcal D_{F,k})-\nabla_{XY}^2G(X_k,Y_k^D;\mathcal D_{G,k})\\
&\quad\times\Biggl[\eta\sum_{i=0}^{Q}\left\{\prod_{j=i+1}^{Q}\left(I-\eta\nabla_{YY}^2G(X_k,Y_k^D;\mathcal B_{k,j})\right)\right\}\nabla_YF(X_k,Y_k^D;\mathcal D_{F,k})\Biggr].
\end{aligned}
\end{equation}
The estimator in~\eqref{eq:stochastic_hg} is evaluated by setting $u_Q=\nabla_YF(X_k,Y_k^D;\mathcal D_{F,k})$, $u_{i-1}=(I-\eta\nabla_{YY}^2G(X_k,Y_k^D;\mathcal B_{k,i}))[u_i]$, and $v_Q=\eta\sum_{i=0}^{Q}u_i$, followed by $\widehat{\nabla\Phi}(X_k)=\nabla_XF(X_k,Y_k^D;\mathcal D_{F,k})-\nabla_{XY}^2G(X_k,Y_k^D;\mathcal D_{G,k})[v_Q]$.
This requires $Q$ Hessian--vector products without forming the Hessian
inverse; see Algorithm~\ref{alg:neumann} in Appendix~\ref{app:algorithm-details}.

Next, we aggregate this estimator using exponential moving average (EMA) momentum:
\begin{equation}\label{eq:momentum_update}
M_k=\theta M_{k-1}+(1-\theta)\widehat{\nabla\Phi}(X_k),\qquad M_{-1}=0,
\end{equation}
where $\theta\in[0,1)$ is the momentum coefficient.
The UL variable $X_k$ is updated by an orthogonalized step: $$X_{k+1}=X_k-\beta O_k,$$ where $\beta>0$ is the UL step-size.
Algorithm~\ref{alg:bimuon} specifies the Exact-Polar direction in Option I and the Finite-NS direction in Option II.

\section{Finite-Time Convergence Rate Analysis of \alg}\label{sec:theory}
In this section, we first establish the convergence of the Exact-Polar \alg version and then extend the analysis to the Finite-NS \alg version.

\subsection{Assumptions}

Toward this end, we begin with several technical assumptions that are useful in our subsequent theoretical analysis.
\begin{assumption}[Strongly Convex LL Problem]\label{ass:convex}
The function $g(X,\cdot)$ is $\mu$-strongly convex, with $\mu>0$.
The same property holds for $G(X,\cdot;\zeta)$ almost surely.
The function $\Phi$ may be nonconvex and satisfies $\Phi_{\inf}:=\inf_X\Phi(X)>-\infty$.
\end{assumption}
\begin{assumption}[Lipschitz Continuity and Smoothness]\label{ass:smooth}
The functions $f,F$ are continuously differentiable and $g,G$ are twice continuously differentiable.
There are constants ${C_f},L>0$ and $\tau,\rho\ge0$ such that, for all $(X,Y),(X',Y')$ and almost every sample,
$
|F(X,Y;\xi)-F(X',Y';\xi)|\le {C_f}\|(X-X',Y-Y')\|_\F,$
$\|\nabla F(X,Y;\xi)-\nabla F(X',Y';\xi)\|_\F\le L\|(X-X',Y-Y')\|_\F,$
$\|\nabla G(X,Y;\zeta)-\nabla G(X',Y';\zeta)\|_\F\le L\|(X-X',Y-Y')\|_\F,$
$\|\nabla_{XY}^2G(X,Y;\zeta)-\nabla_{XY}^2G(X',Y';\zeta)\|_{\op}
\le\tau\|(X-X',Y-Y')\|_\F,$ and
$\|\nabla_{YY}^2G(X,Y;\zeta)-\nabla_{YY}^2G(X',Y';\zeta)\|_{\op}
\le\rho\|(X-X',Y-Y')\|_\F.$
The same inequalities hold for the population functions.
\end{assumption}
\begin{assumption}[Stochastic Derivatives]\label{ass:noise}
All stochastic derivative oracles are unbiased.
Mini-batches are i.i.d.\ and fresh conditional on the history; distinct batches are mutually independent.
Moreover, $\mathbb{E}_{\zeta}
\|\nabla_Y G(X,Y;\zeta)-\nabla_Y g(X,Y)\|_F^2
\le \sigma^2$.
\end{assumption}
These assumptions imply $\|\nabla_XF\|_\F,\|\nabla_YF\|_\F\le {C_f}$, $\|\nabla_{XY}^2G\|_{\op}\le L$, and $\mu I\preceq\nabla_{YY}^2G\preceq LI$.
The last two inequalities are operator bounds induced by the Frobenius norms.

Throughout this section, $D\ge1$ and ${B_{\mathrm{low}}},{B_f},{B_{xy}},{B_H}$ are fixed positive integers.
Let $L_\Phi$ denote the UL smoothness constant given in Lemma~\ref{lem:smooth} of Appendix~\ref{app:svd}.
Algorithm~\ref{alg:bimuon} returns $X_T$ in practice, while the analysis
uses $X_R$ with $R\sim\operatorname{Unif}\{0,\ldots,T-1\}$ independent of
the algorithmic randomness. Thus,
$
\E\|\nabla\Phi(X_R)\|_*
=
\frac{1}{T}\sum_{k=0}^{T-1}\E\|\nabla\Phi(X_k)\|_*.
$

\subsection{Convergence Results of \alg}
When UL movement shifts the LL target, the resulting tracking error contributes to hypergradient bias and then to EMA error.
The following bound quantifies the first part by UL momentum rank.
\begin{lemma}[Rank-Dependent UL--LL Tracking]\label{lem:joint-tracking}
Under Assumptions~\ref{ass:convex}--\ref{ass:noise},
choosing $0<\alpha\le1/L$, $0<\eta<1/L$, $0\le\theta<1$, $\beta>0$, and $D\ge1$ in Algorithm~\ref{alg:bimuon}, we have
\begin{equation}\label{eq:rank-lower}
\frac1T\sum_{k=0}^{T-1}\E\|Y_k^D-Y^\star(X_k)\|_\F^2
\le \frac{\|Y_0-Y^\star(X_0)\|_\F^2}{\alpha\mu T}
+\frac{2\alpha\sigma^2}{\mu {B_{\mathrm{low}}}}
+\frac{L^2\beta^2}{\alpha^2\mu^4T}
\sum_{k=0}^{T-2}\E\operatorname{rank}(M_k).
\end{equation}
\end{lemma}

For Finite-NS \alg, we measure the finite-NS approximation error relative to the exact polar direction.
For a run of length $T$, suppose that
\begin{equation}\label{eq:epsilon-def}
\|\NS(M_k,q)-\polar(M_k)\|_{\op}\le\varepsilon_q,
\qquad 0\le k<T,
\end{equation}
almost surely for some deterministic $\varepsilon_q\in[0,1)$.
Here, $\varepsilon_q$ denotes the NS approximation error and is distinct from the target stationarity accuracy $\epsilon$.

\begin{lemma}[Finite-NS Approximation; {\citealp{kim2026muon}}]\label{lem:ns-approximation}
For the degree-$d$ Taylor NS iteration in Section~\ref{sec:muon}, suppose that the initial orthogonality residual is uniformly bounded by $\delta_{k,0}\le\delta_0<1$ almost surely for all $0\le k<T$.
Then~\eqref{eq:epsilon-def} holds with $\varepsilon_q\le1-\sqrt{1-\delta_0^{(d+1)^q}}\le\delta_0^{(d+1)^q}$.
Hence, for any fixed run satisfying $\delta_0<1$, the NS approximation
error decays doubly exponentially with $q$.
\end{lemma}
Due to space limitations, the residual definition and additional approximation details are provided in Appendix~\ref{app:ns}.
We now present the convergence guarantee for Exact-Polar and Finite-NS \algn.

\begin{theorem}[Exact-Polar \alg and Finite-NS \alg]\label{thm:main}
Suppose Assumptions~\ref{ass:convex}--\ref{ass:noise} hold, and fix $\beta_0>0$.
For any integer $T\ge2$, run Algorithm~\ref{alg:bimuon} with $\alpha=1/(L\sqrt T)$, $\eta=1/(2L)$, $\theta=1-1/\sqrt T$, $\beta=\beta_0/T^{3/4}$, and $Q=\max\{1,\lceil\log T/(4|\log(1-\eta\mu)|)\rceil\}$.

\textnormal{(i) Exact-Polar \alg:} If $O_k=\polar(M_k)$, then $\E\|\nabla\Phi(X_R)\|_*\le C_{\mathrm{pol}}T^{-1/4}$.

\textnormal{(ii) Finite-NS \alg:} If $O_k=\NS(M_k,q)$ for $q\ge0$ and~\eqref{eq:epsilon-def} holds, then $\E\|\nabla\Phi(X_R)\|_*\le\frac{C_{\mathrm{pol}}}{1-\varepsilon_q}T^{-1/4}$.
Here $C_{\mathrm{pol}}$ is independent of $T$; its explicit expression is given in~\eqref{eq:C-svd} of Appendix~\ref{app:svd}.
The NS rate is uniform in $T$ whenever a common $\varepsilon_q<1$
can be chosen independently of $T$.
\end{theorem}
The proofs of these two cases are provided in Appendices~\ref{app:svd} and~\ref{app:ns}, respectively.
\begin{corollary}[Iteration and Stochastic Derivative Complexity]\label{cor:oracle-complexity}
Under the conditions of Theorem~\ref{thm:main}, with a common $\varepsilon_q<1$ independent of $T$ for Finite-NS \alg, both Exact-Polar \alg and Finite-NS \alg require
$
T=O(\epsilon^{-4})
$
outer iterations to obtain $X_R$ with $\E\|\nabla\Phi(X_R)\|_*\le\epsilon$, our criterion for $\epsilon$-stationarity.
Under the oracle schedule of Theorem~\ref{thm:main}, the corresponding stochastic
derivative complexity is
$\widetilde O(\epsilon^{-4})$.
Detailed derivative counts are given in Appendix~\ref{app:ns}.
\end{corollary}
\begin{remark}[Finite-NS and Exact-Polar]
For each fixed run, Lemma~\ref{lem:ns-approximation} shows that the NS approximation error decreases
doubly exponentially as the depth $q$ increases.
Hence, the finite-NS result approaches that of Exact-Polar rapidly with increasing $q$.
\end{remark}

\begin{remark}
The resulting $O(T^{-1/4})$ convergence rate {\em matches} the best-known rate for stochastic single-level Muon-type optimization under matrix parameterizations~\citep{kim2026muon}.
Thus, despite the additional LL tracking and
hypergradient estimation errors introduced by the bilevel structure, \alg preserves the convergence order of its single-level counterparts.
Here, stationarity is measured in the nuclear norm, consistent with the spectral geometry of the polar update.
\end{remark}

\section{\alg with Blockwise Diagonal Matrices and Relative Step-Size Allocation}
\label{sec:blockwise}
\label{sec:multi_block}

To improve the computational and parallel efficiency of \alg, we consider \alg with blockwise diagonal matrices (called \emph{Blockwise Matrix \alg} for convenience subsequently), where the Muon-style orthogonalized direction can be computed separately in parallel for each UL matrix block, followed by a block-specific relative stepsize.
Note that modern neural networks contain many specially structured matrix-valued parameter blocks that are {\em weakly or sparsely} coupled through the network.
Jointly processing all matrix parameters can be expensive, whereas blockwise processing reduces matrix sizes and supports parallel computation
\citep{zhang2017block,mendlerdunner2020randomized,bahamou2023mini} without incurring high inaccuracy.
Next, we restate a theoretical result from \citet{amari2019fisher} to motivate our blockwise design.

\begin{example}[Motivating Example for Blockwise Processing]
{\em Under the wide random-network assumptions of \citet{amari2019fisher}, the Fisher information matrix is approximately block diagonal, with
off-block terms of order $O_p(n^{-1/2})$, where $n$ is the hidden-layer width.
Thus, in this regime, cross-block interactions in the Fisher geometry become weak as the network width increases.}
\end{example}
It is worth noting that the blockwise diagonal matrix setting is also consistent with using Muon for LLM training: in Transformer training, applying Muon separately to the query, key, and value matrices has been reported to perform better than treating a fused QKV projection as a single matrix~\citep{jordan2024muon}.

We write the UL variable as $X=(X^{(1)},\ldots,X^{(J)})$, where $X^{(b)}\in\mathbb{R}^{m_b\times n_b}$ denotes the $b$th matrix block.
Let $M_k^{(b)}$ and $O_k^{(b)}$ denote its hypergradient momentum and Muon-style orthogonalized direction, respectively.
The UL updates of Blockwise Matrix \alg can be written as
\begin{equation}\label{eq:block-update}
X_{k+1}^{(b)}=X_k^{(b)}-\beta\omega_{b,k}O_k^{(b)},
\qquad \omega_{b,k}\geq0,
\end{equation}
where $\beta>0$ is the global UL stepsize and $\omega_{b,k}$ determines the relative stepsize of block $b$.
Conceptually, the block momenta may be organized as $\overline M_k=\operatorname{blkdiag}(M_k^{(1)},\ldots,M_k^{(J)})$, so that the Muon directions are computed on smaller matrices in parallel.

Importantly, the blockwise implementation does not require the UL matrix blocks to be independent.
Instead, we control their simultaneous effect on the moving LL target.
Suppose that, for constants $L_{XY,b}>0$, $\|\nabla_Y g(X,Y)-\nabla_Y g(X',Y)\|_\F^2 \leq\sum_{b=1}^J L_{XY,b}^2\|X^{(b)}-X'^{(b)}\|_\F^2$.
This is a joint bound on simultaneous changes of all matrix blocks rather than an independence assumption.
Let $r_b:=\min\{m_b,n_b\}$ and define the blockwise LL sensitivity
$
w_b:=\frac{L_{XY,b}^2r_b}{\mu^2}.
$
Since $\|O_k^{(b)}\|_\F^2\leq r_b$, the $\mu$-strong convexity of the LL problem yields $\|Y^\star(X_{k+1})-Y^\star(X_k)\|_\F^2 \leq\beta^2\sum_{b=1}^J w_b\omega_{b,k}^2$.
Hence, $w_b$ quantifies how strongly the movement of block $b$ can affect the LL target while remaining valid under joint block updates.
In particular, blocks with larger $w_b$ should be updated more conservatively.

The LL sensitivity alone, however, should not determine the relative stepsize, since a sensitive block may also provide a strong UL signal.
We therefore define
$
a_{b,k}:=\langle M_k^{(b)},O_k^{(b)}\rangle
$
as the predicted UL progress associated with block $b$, and choose the relative stepsizes through
\begin{equation}\label{eq:block-drift-budget}
\max_{\omega_{1,k},\ldots,\omega_{J,k}\geq0}
\sum_{b=1}^J a_{b,k}\omega_{b,k}
\qquad\mathrm{s.t.}\qquad
\sum_{b=1}^J w_b\omega_{b,k}^2\leq1.
\end{equation}
For nonnegative scores that are not all zero, weighted Cauchy--Schwarz yields
\begin{equation}\label{eq:block-allocation-solution}
\omega_{b,k}=\frac{a_{b,k}/w_b}{\sqrt{\sum_{c=1}^J a_{c,k}^2/w_c}}.
\end{equation}
Thus, blocks with stronger predicted UL progress receive larger relative stepsizes, whereas blocks with greater influence on the LL target are more strongly regulated.
Blockwise Matrix \alg therefore retains the computational and parallel advantages of separate Muon-style orthogonalization while using the bilevel structure to guide relative stepsize allocation.

\section{Experimental Results}\label{sec:experiments}
We evaluate \alg on CIFAR-10 representation learning, {CIFAR-100 Transformer representation learning,} and CIFAR-100 blockwise relative stepsize allocation with controlled block sensitivities.
{Appendix~\ref{app:controlled_ul_shapes} further isolates the effects of the UL condition number and spectrum shape.}

\subsection{CIFAR-10 Representation Learning}
\label{sec:real_data}
We evaluate \alg on a CIFAR-10~\citep{krizhevsky2009learning} classification task using frozen pretrained ResNet-18~\citep{he2016deep} features and a matrix-valued UL representation $X\in\mathbb{R}^{256\times512}$.

\textbf{1) Setup.}
The LL variable is a regularized linear classifier trained on the training split, while the UL objective is cross-entropy on a separate validation split.
All methods use the same stocBiO~\citep{ji2021bilevel} LL solver and stochastic implicit hypergradient estimator.
{The five EMA-based methods use the momentum in~\eqref{eq:momentum_update}.}
Here, optimized UL validation loss denotes validation cross-entropy after independently refitting the LL classifier at each logged $X_k$.
Full data splits, solver parameters and tuning details are provided in Appendix~\ref{app:real_data}.

\textbf{2) Baselines.}
We compare {six} UL update rules.
\textbf{stocBiO+EMA} uses the hypergradient momentum $M_k$ directly, while \textbf{OpNorm} and \textbf{FrobNorm} divide $M_k$ by its operator and Frobenius norms, respectively.
Both scalar normalizations change the overall update magnitude while preserving the relative singular spectrum.
{\textbf{stocBiO+Adam} instead applies Adam~\citep{kingma2015adam} directly to the stochastic hypergradient, maintaining its own first- and second-moment estimates.}
Our proposed \textbf{exact-polar \alg} uses the supported polar factor.
Its finite-NS variant (called \textbf{NS-\alg} for convenience) approximates this transformation using 12 NS iterations.
All methods share the same stochastic derivative budget and use stepsizes selected under the same tuning protocol.

\begin{figure}[t!]
\setlength{\parskip}{0pt}
\setlength{\abovecaptionskip}{4pt}
\centering
\begin{minipage}[t]{.332\linewidth}
\centering
\includegraphics[width=\linewidth,trim=0 6pt 0 1pt,clip]{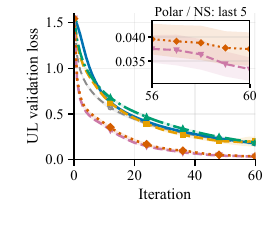}
\par\small(a) UL optimization comparison
\end{minipage}\hfill
\begin{minipage}[t]{.332\linewidth}
\centering
\includegraphics[width=\linewidth,trim=0 6pt 0 1pt,clip]{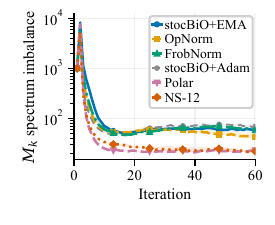}
\par\small(b) Momentum spectrum diagnostic
\end{minipage}\hfill
\begin{minipage}[t]{.333\linewidth}
\centering
\includegraphics[width=\linewidth,trim=0 6pt 0 1pt,clip]{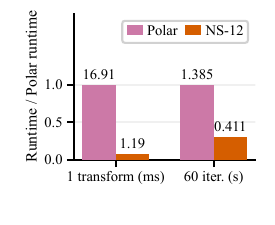}
\par\small(c) Polar vs. NS runtime
\end{minipage}
\caption{CIFAR-10 representation learning and A100 runtime comparison.}
\label{fig:realdata_ul}
\label{fig:realdata_momentum_spectrum}
\label{fig:gpu_timing}
\end{figure}

\textbf{3) Results.}
Figure~\ref{fig:realdata_ul}(a) shows that Exact-Polar \alg and Finite-NS \alg achieve substantially lower optimized UL validation loss than stocBiO+EMA
and the scalar-normalized baselines, supporting spectral reshaping beyond global rescaling.
{Both variants also outperform stocBiO+Adam, the strongest baseline (Table~\ref{tab:realdata_final} in Appendix~\ref{app:real_data}).}
Figure~\ref{fig:realdata_momentum_spectrum}(b) shows lower spectral imbalance
of the pre-transformation momentum $M_k$ under both \alg variants; the metric
is the P90/P10 singular-value ratio defined in Appendix~\ref{app:real_data}.
Figure~\ref{fig:gpu_timing}(c) shows that NS-12 is $14.28\times$ faster than
Exact-Polar per UL transformation and $3.38\times$ faster over 60 updates on an A100 GPU.

{
\subsection{CIFAR-100 Transformer Representation Learning}
\label{sec:transformer_main}
\begin{wrapfigure}{r}{.666\textwidth}
\setlength{\parskip}{0pt}
\setlength{\abovecaptionskip}{3pt}
\centering
\begin{minipage}[t]{.495\linewidth}
\centering
\includegraphics[width=\linewidth]{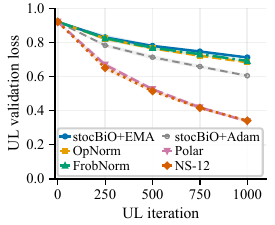}
\par\small(a) UL loss vs. iteration
\end{minipage}\hfill
\begin{minipage}[t]{.495\linewidth}
\centering
\includegraphics[width=\linewidth]{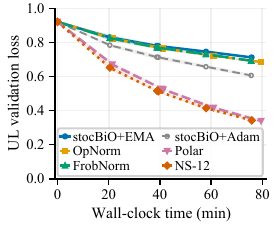}
\par\small(b) UL loss vs. wall-clock time
\end{minipage}
\caption{CIFAR-100 Transformer representation learning and H100 runtime comparison.}
\label{fig:transformer_progress}
\end{wrapfigure}
\textbf{1) Setup.}
We further evaluate \alg at a larger scale on CIFAR-100~\citep{krizhevsky2009learning} with an ImageNet-pretrained DeiT-S/16 Transformer~\citep{touvron2021deit}.
The UL variables are the 16 attention and MLP weight matrices of the last four Transformer blocks (7.1M parameters, 54 times the UL variable in Section~\ref{sec:real_data}), the LL variable is a linear classifier on the CLS features, and each run uses 1,000 UL iterations.
We compare the same six UL updates as in Section~\ref{sec:real_data}, with stepsizes selected under a common protocol.
Full experimental and runtime details are provided in Appendix~\ref{app:transformer}.

\begin{table}[t!]
\centering\small
\setlength{\tabcolsep}{4pt}

\begin{tabular}{lrrr}
\toprule
Method & Final UL loss & Time to $0.75$ (min) & Time to $0.60$ (min)\\
\midrule
stocBiO+EMA & $0.7116\pm0.0040$ & $61.3\pm8.1$ & not reached\\
OpNorm & $0.6835\pm0.0046$ & $60.7\pm0.3$ & not reached\\
FrobNorm & $0.6914\pm0.0050$ & $57.7\pm0.3$ & not reached\\
stocBiO+Adam & $0.6044\pm0.0085$ & $39.1\pm0.2$ & 2/5 seeds\\
Polar & $\mathbf{0.3368\pm0.0062}$ & $21.6\pm0.1$ & $41.1\pm0.2$\\
NS-12 & $0.3412\pm0.0053$ & $\mathbf{20.6\pm0.1}$ & $\mathbf{39.2\pm0.2}$\\
\bottomrule
\end{tabular}
\caption{Transformer results with frozen stepsizes: final UL validation loss after 1,000 iterations, and wall-clock minutes from the start of the run to the first evaluation with UL loss at most $0.75$ and $0.60$. Means and sample standard deviations over five fresh paired seeds.}
\label{tab:transformer_results}
\end{table}

\textbf{2) Results.}
Table~\ref{tab:transformer_results} and Figure~\ref{fig:transformer_progress} show that both \alg variants more than halve the final UL validation loss of stocBiO+EMA and are lower than every baseline, including stocBiO+Adam, on all five seeds.
NS-\alg reaches a UL loss of $0.75$ $3.0\times$ faster in wall-clock time than stocBiO+EMA and $1.9\times$ faster than stocBiO+Adam, while its per-iteration time is within $0.2\%$ of that of stocBiO+EMA.
On the H100, NS-12 reduces the UL-update time from $253.4\pm1.5$ ms for exact polar to $12.3\pm0.1$ ms, a $20.6\times$ speedup for this component.
}

\Needspace{6\baselineskip}
\subsection{Blockwise Stepsize Allocation on CIFAR-100}
\label{sec:cifar100_blockwise}
\begin{wrapfigure}{R}{.666\textwidth}
\setlength{\parskip}{0pt}
\vspace{-.15in}
\centering
\begin{minipage}[t]{.495\linewidth}
\centering
\includegraphics[width=\linewidth,trim=0 2pt 0 3pt,clip]{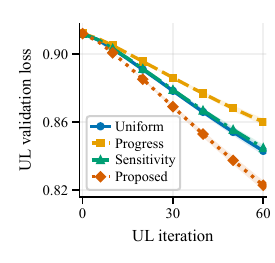}
\par\small (a) $\varrho=10$ trajectory
\end{minipage}\hspace{.009\linewidth}%
\begin{minipage}[t]{.495\linewidth}
\centering
\includegraphics[width=\linewidth,trim=0 2pt 0 3pt,clip]{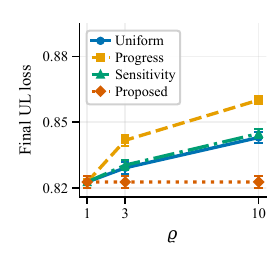}
\par\small (b) Final UL loss vs. $\varrho$
\end{minipage}
\caption{CIFAR-100 blockwise allocation.}
\label{fig:cifar100_blockwise}
\end{wrapfigure}
We further evaluate the blockwise relative-stepsize allocation in Section~\ref{sec:blockwise} on CIFAR-100~\citep{krizhevsky2009learning}.

\textbf{1) Setup.}
We use frozen image features and partition the LL classifier into four matrix blocks.
To obtain controlled and exactly known block sensitivities, block $b$ is coupled to its UL matrix through a coefficient $c_b$, yielding $w_b=L_{XY,b}^2r_b/\mu^2=c_b^2r_b$.
We vary the sensitivity heterogeneity through $\varrho$ and evaluate $\varrho\in\{1,3,10\}$.
Here, $\varrho=\max_b c_b/\min_b c_b$ is the sensitivity ratio, distinct from the Hessian-Lipschitz constant $\rho$ in Assumption~\ref{ass:smooth}.
All compared allocations use the same bilevel estimator, blockwise update-direction construction, global stepsize, and theoretical LL-drift budget $\sum_b w_b\omega_{b,k}^2=1$.
Full construction details are given in Appendix~\ref{app:cifar100_blockwise}.

\textbf{2) Baselines.}
We compare four relative-stepsize rules: \textbf{Uniform}, which ignores block differences; \textbf{Progress-only}, based on the UL score $a_{b,k}$; \textbf{Sensitivity-only}, based on $1/w_b$; and the \textbf{Proposed} allocation, based on $a_{b,k}/w_b$.
Thus, the comparison isolates whether UL progress and LL sensitivity should be considered jointly when distributing the same blockwise movement budget.

\textbf{3) Results.}
Figure~\ref{fig:cifar100_blockwise} shows that the proposed allocation achieves the best UL optimization among the tested rules.
For both $\varrho=3$ and $\varrho=10$, it outperforms Uniform, Progress-only, and Sensitivity-only on all five evaluation seeds.
The advantage becomes clearer when block sensitivities are more heterogeneous, supporting the use of both $a_{b,k}$ and $w_b$ in the relative-stepsize rule.

\section{Conclusion}
In this paper,
we proposed a Muon-type approach called \alg for bilevel optimization and established
$O(T^{-1/4})$ expected nuclear-norm stationarity convergence and
$\widetilde{O}(\epsilon^{-4})$ stochastic derivative complexity guarantees for its exact-polar and finite-NS variants.
For increased computational efficiency, we further investigated \alg with blockwise diagonal approximation and derived a bilevel-informed relative
step-size allocation rule that balances UL progress and LL sensitivity across matrix blocks.
Experiments on representation learning and blockwise bilevel optimization problems verified the efficacy of our proposed \alg designs.
\label{main:end}
\bibliography{references}
\bibliographystyle{references}
\clearpage
\appendix
\raggedbottom
\setlength{\textfloatsep}{10pt plus 2pt minus 2pt}
\setlength{\floatsep}{8pt plus 2pt minus 2pt}
\setlength{\abovecaptionskip}{4pt}
\section{Derivations and Details for Bilevel-Induced Spectral Imbalance}
\label{app:quadratic}
This appendix gives the derivations supporting Section~\ref{sec:bilevel-spectral-imbalance}.
We first derive the LL solution and induced UL curvature, then make the
aligned hypergradient calculation explicit and describe the controlled
numerical construction used in Figures~\ref{fig:spectral-illustration} and~\ref{fig:spectral-supplement}.

\subsection{LL solution and induced UL curvature}
For the model in Section~\ref{sec:bilevel-spectral-imbalance}, let $Z_{\mathrm{tr}}\in\mathbb R^{n_{\mathrm{tr}}\times m}$, $Z_{\mathrm{val}}\in\mathbb R^{n_{\mathrm{val}}\times m}$, $R_{\mathrm{tr}}\in\mathbb R^{n_{\mathrm{tr}}\times n}$, and $R_{\mathrm{val}}\in\mathbb R^{n_{\mathrm{val}}\times n}$, with positive sample counts.
The matrices $X,Y,X_{\mathrm{prior}}$ lie in $\mathbb R^{m\times n}$, and $\mu>0$, $\gamma\ge0$.
Using the notation of the main text, let $H_{\mathrm{LL}}=\frac{Z_{\mathrm{tr}}^\top Z_{\mathrm{tr}}}{n_{\mathrm{tr}}}+\mu I$, $\Sigma_{\mathrm{val}}=\frac{Z_{\mathrm{val}}^\top Z_{\mathrm{val}}}{n_{\mathrm{val}}}$, and $H_{\mathrm{LL}}=P\operatorname{diag}(h_1,\ldots,h_m)P^\top$, where $P$ is orthogonal and $h_i\ge\mu>0$.

\paragraph{LL solution and sensitivity.}
Direct differentiation gives $\nabla_Yg(X,Y)=H_{\mathrm{LL}}Y-\frac{Z_{\mathrm{tr}}^\top R_{\mathrm{tr}}}{n_{\mathrm{tr}}}-\mu X$, and $\nabla_{YY}^2g(X,Y)[\Delta Y]=H_{\mathrm{LL}}\Delta Y$.
Since $H_{\mathrm{LL}}\succeq\mu I\succ0$, the unique LL solution is
\begin{equation}\label{eq:quad_ystar}
Y^\star(X)=H_{\mathrm{LL}}^{-1}\left(\frac{Z_{\mathrm{tr}}^\top R_{\mathrm{tr}}}{n_{\mathrm{tr}}}+\mu X\right).
\end{equation}
Differentiating this affine map gives $DY^\star(X)[\Delta X]=\mu H_{\mathrm{LL}}^{-1}\Delta X$.
The inverse acts on the left; the perturbations and solution have shape $m\times n$.

\paragraph{UL derivatives.}
The partial derivatives are $\nabla_X f=\gamma(X-X_{\mathrm{prior}})$ and $\nabla_Y f=\Sigma_{\mathrm{val}}Y-Z_{\mathrm{val}}^\top R_{\mathrm{val}}/n_{\mathrm{val}}$.
The chain rule and symmetry of $H_{\mathrm{LL}}^{-1}$ yield
\begin{equation}\label{eq:proximal-gradient}
\nabla\Phi(X)=\gamma(X-X_{\mathrm{prior}})+\mu H_{\mathrm{LL}}^{-1}\left(\Sigma_{\mathrm{val}}Y^\star(X)-\frac{Z_{\mathrm{val}}^\top R_{\mathrm{val}}}{n_{\mathrm{val}}}\right).
\end{equation}
Differentiating once more gives $\mathcal H_{\mathrm{UL}}[\Delta X] =\nabla^2\Phi(X)[\Delta X] =\left(\mu^2H_{\mathrm{LL}}^{-1}\Sigma_{\mathrm{val}}H_{\mathrm{LL}}^{-1}+\gamma I\right)\Delta X$.
The operator acts identically and independently on each column.
The target terms and prior affect the gradient and optimum but not this constant Hessian.
For general $\Sigma_{\mathrm{val}}\succeq0$, the Hessian is positive semidefinite; it is positive definite if $\Sigma_{\mathrm{val}}\succ0$ or $\gamma>0$.
No finite condition number is claimed in a singular case.

\paragraph{Squared conditioning under isotropic validation.}
When $\Sigma_{\mathrm{val}}=I$, the matrices $Pe_ie_j^\top$, $1\le i\le m$, $1\le j\le n$, form an orthonormal Frobenius-space eigenbasis of $\mathcal H_{\mathrm{UL}}$ with eigenvalues $\mu^2/h_i^2+\gamma$.
Here $e_i,e_j$ are the standard coordinate vectors in the respective spaces.
Writing $h_{\min}=\min_i h_i$ and $h_{\max}=\max_i h_i$ gives $\kappa(\mathcal H_{\mathrm{UL}}) =\frac{\gamma+\mu^2/h_{\min}^2}{\gamma+\mu^2/h_{\max}^2}$.
In particular, when $\gamma=0$, $\mathcal H_{\mathrm{UL}}[\Delta X]=\mu^2H_{\mathrm{LL}}^{-2}\Delta X$, $\kappa(\mathcal H_{\mathrm{UL}}) =\left(\frac{h_{\max}}{h_{\min}}\right)^2 =\kappa(H_{\mathrm{LL}})^2$.
This is the identity stated in Section~\ref{sec:bilevel-spectral-imbalance}.
The common factor $\mu^2$ cancels from the condition number, whereas positive $\gamma$ moves the ratio toward one.
More generally, if $(1-\delta)cI\preceq\Sigma_{\mathrm{val}}\preceq(1+\delta)cI$ for $c>0$ and $0\le\delta<1$, congruence gives $(1-\delta)cH_{\mathrm{LL}}^{-2} \preceq H_{\mathrm{LL}}^{-1}\Sigma_{\mathrm{val}}H_{\mathrm{LL}}^{-1} \preceq(1+\delta)cH_{\mathrm{LL}}^{-2}$.
Thus approximate isotropy gives spectral bounds, not exact equality or exact polar invariance.
\subsection{Aligned Hypergradient Singular Magnitudes and Scope}

Under $\Sigma_{\mathrm{val}}=I$ and $\gamma=0$, define $X^\star = \mu^{-1} \left( H_{\mathrm{LL}} \frac{Z_{\mathrm{val}}^\top R_{\mathrm{val}}}{n_{\mathrm{val}}} - \frac{Z_{\mathrm{tr}}^\top R_{\mathrm{tr}}}{n_{\mathrm{tr}}} \right)$.
Substituting~\eqref{eq:quad_ystar} into the UL gradient gives $\nabla\Phi(X) = \mu^2 H_{\mathrm{LL}}^{-2}(X-X^\star)$.
Let $r=\min\{m,n\}$, let $P_r$ contain $r$ eigenvectors of
$H_{\mathrm{LL}}$, and let $V\in\mathbb{R}^{n\times r}$ satisfy
$V^\top V=I_r$.
For an aligned displacement $X-X^\star = P_r\operatorname{diag}(x_1,\ldots,x_r)V^\top$, $x_i>0$, we obtain $\nabla\Phi(X) = \mu^2 P_r \operatorname{diag} \left( \frac{x_1}{h_1^2},\ldots,\frac{x_r}{h_r^2} \right) V^\top$.
Hence, the hypergradient singular magnitudes scale as
$\mu^2x_i/h_i^2$, showing explicitly how LL curvature can induce
spectral imbalance in the UL hypergradient.
Moreover, $\operatorname{polar}(\nabla\Phi(X))=P_rV^\top$, so the exact polar direction removes the positive singular-value scaling
while preserving the active singular subspaces.

These identities rely on the quadratic model, isotropic validation
curvature, $\gamma=0$, and the aligned displacement above.
The example therefore provides a local second-order mechanism for the
spectral effect discussed in Section~\ref{sec:bilevel-spectral-imbalance}, rather than a general identity
for arbitrary smooth bilevel problems.

\subsection{Controlled Numerical Illustration of the
Section~\ref{sec:bilevel-spectral-imbalance} Quadratic Model}
\label{app:controlled-spectral}
\begin{figure}[!htbp]
\centering
\begin{minipage}[t]{.42\linewidth}
\centering
\includegraphics[width=\linewidth]{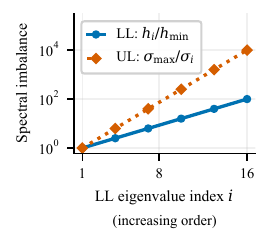}
\par\small (a) Spectral amplification
\end{minipage}\hfill
\begin{minipage}[t]{.42\linewidth}
\centering
\includegraphics[width=\linewidth]{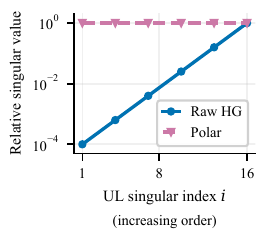}
\par\small (b) Polar normalization
\end{minipage}
\caption{Additional illustrations of the quadratic mechanism in~\eqref{eq:quadratic-motivation}.
(a) For aligned $x_i=1$, a $10^2$ LL Hessian spectral spread induces a $10^4$ UL hypergradient singular-value spread.
(b) The supported polar factor of the same hypergradient has unit singular values; raw singular values are normalized by their maximum.}
\label{fig:spectral-supplement}
\end{figure}
We use the quadratic model in
Section~\ref{sec:bilevel-spectral-imbalance} with
$m=n=n_{\mathrm{tr}}=n_{\mathrm{val}}=16$,
$\mu=0.1$, $\gamma=0$, and
$R_{\mathrm{tr}}=R_{\mathrm{val}}=X_{\mathrm{prior}}=0$.
We set $Z_{\mathrm{val}}=\sqrt{n_{\mathrm{val}}}I$, so that
$Z_{\mathrm{val}}^\top Z_{\mathrm{val}}/n_{\mathrm{val}}=I$.

For each prescribed condition number
$\kappa\in\{1,3,10,30,100,300\}$,
we generate an orthogonal matrix $P$ and choose geometrically spaced
eigenvalues $h_i\in[1,\kappa]$.
We then define $H_{\mathrm{LL}} = P\operatorname{diag}(h_1,\ldots,h_m)P^\top$, $Z_{\mathrm{tr}} = \sqrt{n_{\mathrm{tr}}}\, \operatorname{diag}\!\bigl(\sqrt{h_i-\mu}\bigr)P^\top$, which gives
$Z_{\mathrm{tr}}^\top Z_{\mathrm{tr}}/n_{\mathrm{tr}}+\mu I
=H_{\mathrm{LL}}$.

Since the targets are zero, $X^\star=0$.
We use the aligned displacement $X-X^\star = P\operatorname{diag}(x_1,\ldots,x_m)V^\top$, with $x_i=1$ and orthogonal $V$.
Figure~\ref{fig:spectral-illustration} reports the resulting relation
$\kappa(H_{\mathrm{UL}})=\kappa(H_{\mathrm{LL}})^2$.
Figure~\ref{fig:spectral-supplement} uses the $\kappa=100$ instance with seed~0 to visualize the
corresponding hypergradient spectral amplification and exact-polar
normalization.

For numerical verification, we repeat each condition number over five
independent orthogonal rotations using seeds 0--4.
All computations are performed in float64 on CPU.

\section{Algorithm Implementation Details}
\label{app:implementation}
\label{app:algorithm-details}

This appendix gives the finite Newton--Schulz (NS) map and stochastic
Neumann recursion used in Section~\ref{sec:algorithm}.

\subsection{Newton--Schulz Orthogonalization}
\label{app:ns-definition}

Following \citet{kim2026muon}, for an integer $d\geq 1$, define the
degree-$d$ Taylor polynomial of $x^{-1/2}$ around $x=1$ as $p(x) = \sum_{\ell=0}^{d} \frac{(2\ell)!}{4^\ell(\ell!)^2}(1-x)^\ell$.
For a square matrix $A$, its matrix evaluation is $p(A) = \sum_{\ell=0}^{d} \frac{(2\ell)!}{4^\ell(\ell!)^2}(I-A)^\ell$.
Given $M\in\mathbb{R}^{m\times n}$, Algorithm~\ref{alg:ns}
defines the finite-NS map $\operatorname{NS}(M,q)$ used in
Section~\ref{sec:muon}.

\begin{algorithm}[!htbp]
\algcaption{Newton--Schulz orthogonalization}
\label{alg:ns}
\begin{algorithmic}[1]
\Require Matrix $M\in\mathbb{R}^{m\times n}$; integer NS depth $q\ge0$;
degree-$d$ polynomial $p$
\State $O^{[0]}\gets M/\max\{1,\|M\|_\F\}$
\For{$j=0,\ldots,q-1$}
    \State
    $O^{[j+1]}
    \gets
    p\!\left(O^{[j]}(O^{[j]})^\top\right)O^{[j]}$
\EndFor
\State \Return $\operatorname{NS}(M,q)=O^{[q]}$
\end{algorithmic}
\end{algorithm}

By construction,
$\operatorname{NS}(M,0)=O^{[0]}$ and
$\operatorname{NS}(0,q)=0$.

\subsection{Stochastic Neumann Hypergradient Estimation}
\label{app:neumann}

The truncated Neumann term in~\eqref{eq:stochastic_hg}
is evaluated through the Hessian--vector recursion in
Algorithm~\ref{alg:neumann}.
For fixed $(X_k,Y_k^D)$, it uses $Q$ independent Hessian mini-batches
and requires $Q$ Hessian--vector products without explicitly forming
the LL Hessian inverse.

\begin{algorithm}[!htbp]
\algcaption{Stochastic inverse-Hessian estimate}
\label{alg:neumann}
\begin{algorithmic}[1]
\Require Fixed $(X_k,Y_k^D)$; stepsize $\eta$; depth $Q$;
loss batch $\mathcal D_{F,k}$; independent Hessian batches
$\{\mathcal B_{k,i}\}_{i=1}^{Q}$, each of size $B_H$
\State
$u_Q\gets\nabla_YF(X_k,Y_k^D;\mathcal D_{F,k})$
\For{$i=Q,\ldots,1$}
    \State
    $u_{i-1}
    \gets
    u_i
    -
    \eta\nabla_{YY}^2G(X_k,Y_k^D;\mathcal B_{k,i})[u_i]$
\EndFor
\State
\Return
$v_Q=\eta\sum_{i=0}^{Q}u_i$
\end{algorithmic}
\end{algorithm}

Using an independent mixed-derivative batch $\mathcal D_{G,k}$, the resulting
stochastic hypergradient estimator is $\widehat{\nabla}\Phi(X_k) = \nabla_XF(X_k,Y_k^D;\mathcal D_{F,k}) - \nabla_{XY}^2G(X_k,Y_k^D;\mathcal D_{G,k})[v_Q]$, which is the estimator in~\eqref{eq:stochastic_hg}.

\section{Convergence proof for exact-polar \alg}\label{app:svd}
This appendix proves the exact-polar result in Theorem~\ref{thm:main}; the same
bilevel tracking bounds are reused for the finite-NS analysis in Appendix~\ref{app:ns}.
Throughout this section, suppose Assumptions~\ref{ass:convex}--\ref{ass:noise} hold.
The Neumann estimator is specified in Appendix~\ref{app:algorithm-details}.

\begin{lemma}[Smoothness of the UL objective]\label{lem:smooth}
Under Assumptions~\ref{ass:convex}--\ref{ass:noise}, the UL objective $\Phi$ is $L_\Phi$-smooth: for all $X,X'\in\mathbb R^{m\times n}$,
\begin{equation}\label{eq:gradLip}
\|\nabla\Phi(X)-\nabla\Phi(X')\|_\F\le L_\Phi\|X-X'\|_\F,
\end{equation}
where $L_\Phi=\left(L+\frac{L^2}{\mu}+\frac{{C_f}\tau}{\mu}+\frac{L{C_f}\rho}{\mu^2}\right)\left(1+\frac L\mu\right)$.
\end{lemma}
\begin{proof}
For fixed $X$, strong convexity gives
\[
g(X,Y)\ge g(X,0)+\langle\nabla_Yg(X,0),Y\rangle+\frac\mu2\|Y\|_\F^2
\ge g(X,0)-\|\nabla_Yg(X,0)\|_\F\|Y\|_\F+\frac\mu2\|Y\|_\F^2.
\]
The right side tends to infinity as $\|Y\|_\F\to\infty$.
Continuity therefore gives existence of a minimizer, and strong convexity gives uniqueness.
The first-order condition is $\nabla_Yg(X,Y^\star(X))=0$.
Strong monotonicity at fixed $X$ and the first-order conditions at $X,X'$ imply
\begin{align*}
\mu\|Y^\star(X)-Y^\star(X')\|_\F^2
&\le\langle\nabla_Yg(X,Y^\star(X))-\nabla_Yg(X,Y^\star(X')),Y^\star(X)-Y^\star(X')\rangle\\
&=\langle\nabla_Yg(X',Y^\star(X'))-\nabla_Yg(X,Y^\star(X')),Y^\star(X)-Y^\star(X')\rangle\\
&\le L\|X-X'\|_\F\|Y^\star(X)-Y^\star(X')\|_\F.
\end{align*}
If the LL solutions differ, divide by their distance; otherwise the following bound is immediate:
\begin{equation}\label{eq:Ylip}
\|Y^\star(X)-Y^\star(X')\|_\F\le\frac L\mu\|X-X'\|_\F.
\end{equation}

The map $(X,Y)\mapsto\nabla_Yg(X,Y)$ is continuously differentiable, with invertible derivative $\nabla_{YY}^2g(X,Y)\succeq\mu I$ in its LL argument.
The implicit function theorem gives a continuously differentiable local solution near every $X$.
Uniqueness makes the local solutions agree, so $Y^\star$ is continuously differentiable globally.
Differentiating the LL first-order condition in an arbitrary UL direction $U$ gives
\begin{align*}
0&=D_X(\nabla_Yg)(X,Y^\star(X))[U]
+\nabla_{YY}^2g(X,Y^\star(X))[DY^\star(X)[U]],\\
DY^\star(X)[U]
&=-(\nabla_{YY}^2g(X,Y^\star(X)))^{-1}D_X(\nabla_Yg)(X,Y^\star(X))[U].
\end{align*}
The chain rule gives
\begin{align*}
D\Phi(X)[U]
&=\langle\nabla_Xf(X,Y^\star(X)),U\rangle
+\langle\nabla_Yf(X,Y^\star(X)),DY^\star(X)[U]\rangle\\
&=\langle\nabla_Xf(X,Y^\star(X)),U\rangle\\
&\quad-\Big\langle\nabla_{XY}^2g(X,Y^\star(X))
[(\nabla_{YY}^2g(X,Y^\star(X)))^{-1}\nabla_Yf(X,Y^\star(X))],U\Big\rangle.
\end{align*}
Here the LL Hessian is self-adjoint, and the mixed derivative is the adjoint of $D_X(\nabla_Yg)$.
This is the matrix form of the implicit hypergradient in~\eqref{eq:exact_hypergradient}.
The bounds on the two loss partials, mixed derivative and inverse Hessian also give $\|\nabla\Phi(X)\|_\F\le {C_f}+L{C_f}/\mu$.

For two pairs $(X,Y),(X',Y')$, the inverse identity gives
\begin{align*}
&(\nabla_{YY}^2g(X,Y))^{-1}-(\nabla_{YY}^2g(X',Y'))^{-1}\\
&=(\nabla_{YY}^2g(X,Y))^{-1}
[\nabla_{YY}^2g(X',Y')-\nabla_{YY}^2g(X,Y)](\nabla_{YY}^2g(X',Y'))^{-1},\\
&\|(\nabla_{YY}^2g(X,Y))^{-1}-(\nabla_{YY}^2g(X',Y'))^{-1}\|_{\op}
\le\frac\rho{\mu^2}\|(X-X',Y-Y')\|_\F.
\end{align*}
To bound the difference of the three-factor term $\nabla_{XY}^2g[(\nabla_{YY}^2g)^{-1}\nabla_Yf]$, change the mixed derivative first, then the inverse Hessian, and finally the loss partial gradient.
Their respective norm bounds are $\frac{{C_f}\tau}{\mu}\|(X-X',Y-Y')\|_\F$, $\frac{L{C_f}\rho}{\mu^2}\|(X-X',Y-Y')\|_\F$, $\frac{L^2}{\mu}\|(X-X',Y-Y')\|_\F$.
Adding the $L$-Lipschitz UL partial gradient proves a joint Lipschitz bound with coefficient $L+\frac{L^2}{\mu}+\frac{{C_f}\tau}{\mu}+\frac{L{C_f}\rho}{\mu^2}\le L_\Phi$.
In particular, setting $X'=X$ and $Y'=Y^\star(X)$ yields the estimate used below to control hypergradient bias:
\begin{align}\label{eq:localLip}
&\|\nabla_Xf(X,Y)-\nabla_{XY}^2g(X,Y)[(\nabla_{YY}^2g(X,Y))^{-1}\nabla_Yf(X,Y)]
-\nabla\Phi(X)\|_\F\nonumber\\
&\qquad\le L_\Phi\|Y-Y^\star(X)\|_\F.
\end{align}
Setting $Y=Y^\star(X)$ and $Y'=Y^\star(X')$, and using~\eqref{eq:Ylip}, gives
\begin{align*}
\|\nabla\Phi(X)-\nabla\Phi(X')\|_\F
&\le\left(L+\frac{L^2}{\mu}+\frac{{C_f}\tau}{\mu}+\frac{L{C_f}\rho}{\mu^2}\right)
\bigl(\|X-X'\|_\F+\|Y^\star(X)-Y^\star(X')\|_\F\bigr)\\
&\le\left(L+\frac{L^2}{\mu}+\frac{{C_f}\tau}{\mu}+\frac{L{C_f}\rho}{\mu^2}\right)(1+L/\mu)\|X-X'\|_\F\\
&=L_\Phi\|X-X'\|_\F.
\end{align*}
Finally, the fundamental theorem of calculus and~\eqref{eq:gradLip} imply
\begin{align*}
\Phi(X+U)-\Phi(X)-\langle\nabla\Phi(X),U\rangle
&=\int_0^1\langle\nabla\Phi(X+tU)-\nabla\Phi(X),U\rangle\,dt\\
&\le\int_0^1L_\Phi t\|U\|_\F^2\,dt
=\frac{L_\Phi}{2}\|U\|_\F^2.
\end{align*}
Thus the smoothness conclusion also gives the descent inequality used in both convergence proofs:
\begin{equation}\label{eq:descent}
\Phi(X+U)\le\Phi(X)+\langle\nabla\Phi(X),U\rangle+\frac{L_\Phi}{2}\|U\|_\F^2.
\end{equation}
\end{proof}

The following proposition bounds the truncation bias of the stochastic
Neumann estimate.
\begin{proposition}[Bias of the inverse-Hessian estimate]\label{prop:neumann}
Fix $(X,Y)$ and let $v_Q$ be the estimate returned by Algorithm~\ref{alg:neumann}, evaluated at $(X,Y)$ in place of $(X_k,Y_k^D)$.
Omit $(X,Y)$ from the derivative arguments.
Let $0<\eta<1/L$ and $|\mathcal B_i|={B_H}$.
Then
\begin{equation}\label{eq:vbias}
\|\E v_Q-(\nabla_{YY}^2g)^{-1}\nabla_Yf\|_\F
\le\frac {C_f}\mu(1-\eta\mu)^{Q+1}.
\end{equation}
\end{proposition}
\begin{proof}
Algorithm~\ref{alg:neumann} and $\mu I\preceq\nabla_{YY}^2G\preceq LI$ give $0\prec I-\eta\nabla_{YY}^2G(\mathcal B_i)\preceq(1-\eta\mu)I, \, \|u_{i-1}\|_\F\le(1-\eta\mu)\|u_i\|_\F, \, \|u_i\|_\F\le {C_f}(1-\eta\mu)^{Q-i}$.
Summing these bounds gives the estimate used later for the stochastic hypergradient:
\begin{equation}\label{eq:vbound}
\|v_Q\|_\F\le\eta {C_f}\sum_{i=0}^Q(1-\eta\mu)^i
=\frac {C_f}\mu[1-(1-\eta\mu)^{Q+1}]\le\frac {C_f}\mu
\quad\text{almost surely}.
\end{equation}
Since $u_i$ depends only on $u_Q,\mathcal B_Q,\ldots,\mathcal B_{i+1}$,
\begin{align*}
&\E[u_{i-1}\mid u_Q,\mathcal B_Q,\ldots,\mathcal B_{i+1}]=(I-\eta\nabla_{YY}^2g)u_i,\\
&\E u_i=(I-\eta\nabla_{YY}^2g)^{Q-i}\nabla_Yf,\\
&\E v_Q=\eta\sum_{i=0}^Q(I-\eta\nabla_{YY}^2g)^i\nabla_Yf\\
&=[I-(I-\eta\nabla_{YY}^2g)^{Q+1}](\nabla_{YY}^2g)^{-1}\nabla_Yf,\\
&\E v_Q-(\nabla_{YY}^2g)^{-1}\nabla_Yf=-(I-\eta\nabla_{YY}^2g)^{Q+1}(\nabla_{YY}^2g)^{-1}\nabla_Yf,\\
&\|\E v_Q-(\nabla_{YY}^2g)^{-1}\nabla_Yf\|_\F\le(1-\eta\mu)^{Q+1}\mu^{-1}{C_f}.
\end{align*}
\end{proof}

We next use the bias decomposition of Lemma~7 in \citet{ji2021bilevel}.
We retain the existing $L_\Phi$ as an upper bound on its local Lipschitz coefficient and the $Q+1$ truncation exponent from Proposition~\ref{prop:neumann}.
\begin{lemma}[Conditional hypergradient bias]\label{lem:hg}
Let $0<\eta<1/L$.
Conditioning on $X_k$ and $Y_k^D$, we have
\begin{equation}\label{eq:hgbias}
\|\E\widehat{\nabla\Phi}(X_k)-\nabla\Phi(X_k)\|_\F^2 \le 2L_\Phi^2\|Y_k^D-Y^\star(X_k)\|_\F^2
+\frac{2L^2{C_f}^2}{\mu^2}(1-\eta\mu)^{2Q+2}.
\end{equation}
\end{lemma}
\begin{proof}
Throughout this proof, $X_k$ and $Y_k^D$ are fixed, and $\E$ is over the fresh batches $\mathcal D_\F,\mathcal D_G,\mathcal B_1,\ldots,\mathcal B_Q$.
The outer index $k$ is suppressed on these batches.
All derivatives are evaluated at $(X_k,Y_k^D)$.
By the conditional independence in Assumption~\ref{ass:noise},
\begin{align*}
&\E\widehat{\nabla\Phi}(X_k)
=\nabla_Xf-\nabla_{XY}^2g[\E v_Q],\\
&\E\widehat{\nabla\Phi}(X_k)-\nabla\Phi(X_k)
=\nabla_Xf-\nabla_{XY}^2g[(\nabla_{YY}^2g)^{-1}\nabla_Yf]-\nabla\Phi(X_k)\\
&\quad-\nabla_{XY}^2g[\E v_Q-(\nabla_{YY}^2g)^{-1}\nabla_Yf].
\end{align*}
By~\eqref{eq:localLip} and~\eqref{eq:vbias},
\begin{align*}
&\|\E\widehat{\nabla\Phi}(X_k)-\nabla\Phi(X_k)\|_\F^2\\
&\le{}2\|\nabla_Xf-\nabla_{XY}^2g[(\nabla_{YY}^2g)^{-1}\nabla_Yf]-\nabla\Phi(X_k)\|_\F^2\\
&+2L^2\|\E v_Q-(\nabla_{YY}^2g)^{-1}\nabla_Yf\|_\F^2\\
&\le{}2L_\Phi^2\|Y_k^D-Y^\star(X_k)\|_\F^2
+\frac{2L^2{C_f}^2}{\mu^2}(1-\eta\mu)^{2Q+2}.
\end{align*}
This proves~\eqref{eq:hgbias}.
\end{proof}

\begin{lemma}[Rank-dependent LL-solution tracking]\label{lem:lower}
For Algorithm~\ref{alg:bimuon} with either exact polar or finite NS, $0<\alpha\le1/L$, $\beta>0$, and $D\ge1$,
\begin{equation}\label{eq:lower-rank-uniform}
\frac1T\sum_{k=0}^{T-1}\E\|Y_k^D-Y^\star(X_k)\|_\F^2
\le{}\frac{\|Y_0-Y^\star(X_0)\|_\F^2}{\alpha\mu T}
+\frac{2\alpha\sigma^2}{\mu {B_{\mathrm{low}}}} +\frac{L^2\beta^2r}{\alpha^2\mu^4}.
\end{equation}
Here $r=\min\{m,n\}$ bounds the rank of the UL momentum matrix.
\end{lemma}
\begin{proof}
The fundamental theorem of calculus along the LL segment gives
\begin{align*}
&\nabla_Yg(X_k,Y_k^t)-\nabla_Yg(X_k,Y^\star(X_k))\\
&=\left[\int_0^1\nabla_{YY}^2g\bigl(X_k,Y^\star(X_k)+u(Y_k^t-Y^\star(X_k))\bigr)\,du\right]
[Y_k^t-Y^\star(X_k)].
\end{align*}
The self-adjoint operator in brackets lies between $\mu I$ and $LI$.
Since $0<\alpha\le1/L$, subtracting $\alpha$ times that operator from $I$ gives eigenvalues in $[0,1-\alpha\mu]$.
Using the zero LL gradient at $Y^\star(X_k)$ therefore gives $\|Y_k^t-\alpha\nabla_Yg(X_k,Y_k^t)-Y^\star(X_k)\|_\F^2 \le(1-\alpha\mu)^2\|Y_k^t-Y^\star(X_k)\|_\F^2$.
By freshness and unbiasedness of the LL batch,
\begin{align*}
&\E[\nabla_YG(X_k,Y_k^t;\mathcal S_{k,t})-\nabla_Yg(X_k,Y_k^t)\mid X_k,Y_k^t]=0,\\
&\E[\|\nabla_YG(X_k,Y_k^t;\mathcal S_{k,t})-\nabla_Yg(X_k,Y_k^t)\|_\F^2\mid X_k,Y_k^t]\le\sigma^2/{B_{\mathrm{low}}}.
\end{align*}
The conditional cross term vanishes, and hence
\begin{align*}
&\E[\|Y_k^{t+1}-Y^\star(X_k)\|_\F^2\mid X_k,Y_k^t]\\
&=\|Y_k^t-\alpha\nabla_Yg(X_k,Y_k^t)-Y^\star(X_k)\|_\F^2\\
&\quad+\alpha^2\E[\|\nabla_YG(X_k,Y_k^t;\mathcal S_{k,t})-\nabla_Yg(X_k,Y_k^t)\|_\F^2\mid X_k,Y_k^t]\\
&\le(1-\alpha\mu)^2\|Y_k^t-Y^\star(X_k)\|_\F^2+\frac{\alpha^2\sigma^2}{{B_{\mathrm{low}}}}.
\end{align*}
Iterating and taking expectations,
\begin{align}\label{eq:inner-exact}
&\E\|Y_k^D-Y^\star(X_k)\|_\F^2
\le{}(1-\alpha\mu)^{2D}\E\|Y_k^0-Y^\star(X_k)\|_\F^2\nonumber\\
&+\frac{\alpha^2\sigma^2}{{B_{\mathrm{low}}}}\sum_{t=0}^{D-1}(1-\alpha\mu)^{2t}\nonumber\\
&={}(1-\alpha\mu)^{2D}\E\|Y_k^0-Y^\star(X_k)\|_\F^2
+\frac{\alpha^2\sigma^2[1-(1-\alpha\mu)^{2D}]}{{B_{\mathrm{low}}}[1-(1-\alpha\mu)^2]}.
\end{align}
If $\alpha\mu=1$, then
\[
\frac1T\sum_{k=0}^{T-1}\E\|Y_k^D-Y^\star(X_k)\|_\F^2
\le\frac{\alpha^2\sigma^2}{{B_{\mathrm{low}}}}=\frac{\alpha\sigma^2}{\mu {B_{\mathrm{low}}}}\le\frac{2\alpha\sigma^2}{\mu {B_{\mathrm{low}}}},
\]
which proves both~\eqref{eq:lower-rank-uniform} and~\eqref{eq:loweravg}.
Suppose $0<\alpha\mu<1$.
For $k\ge1$, Young's inequality with weight $[1-(1-\alpha\mu)^D]/(1-\alpha\mu)^D$, warm starting, and~\eqref{eq:Ylip} give
\begin{align*}
&(1-\alpha\mu)^{2D}\E\|Y_k^0-Y^\star(X_k)\|_\F^2\\
&\le(1-\alpha\mu)^D\E\|Y_{k-1}^D-Y^\star(X_{k-1})\|_\F^2
+\frac{(1-\alpha\mu)^{2D}}{1-(1-\alpha\mu)^D}\E\|Y^\star(X_{k-1})-Y^\star(X_k)\|_\F^2\\
&\le(1-\alpha\mu)^D\E\|Y_{k-1}^D-Y^\star(X_{k-1})\|_\F^2
+\frac{(1-\alpha\mu)^{2D}L^2}{[1-(1-\alpha\mu)^D]\mu^2}\E\|X_k-X_{k-1}\|_\F^2.
\end{align*}
Summing~\eqref{eq:inner-exact}, including $Y_0^0=Y_0$, gives
\begin{align*}
&\sum_{k=0}^{T-1}\E\|Y_k^D-Y^\star(X_k)\|_\F^2\\
&\le(1-\alpha\mu)^{2D}\|Y_0-Y^\star(X_0)\|_\F^2
+(1-\alpha\mu)^D\sum_{k=0}^{T-2}\E\|Y_k^D-Y^\star(X_k)\|_\F^2\\
&\quad+\frac{(1-\alpha\mu)^{2D}L^2}{[1-(1-\alpha\mu)^D]\mu^2}
\sum_{k=0}^{T-2}\E\|X_{k+1}-X_k\|_\F^2
+\frac{T\alpha^2\sigma^2[1-(1-\alpha\mu)^{2D}]}{{B_{\mathrm{low}}}[1-(1-\alpha\mu)^2]}.
\end{align*}
Since the omitted error at $T-1$ is nonnegative, rearrangement yields
\begin{align*}
&\frac1T\sum_{k=0}^{T-1}\E\|Y_k^D-Y^\star(X_k)\|_\F^2
\le{}\frac{(1-\alpha\mu)^{2D}\|Y_0-Y^\star(X_0)\|_\F^2}{T[1-(1-\alpha\mu)^D]}\\
&+\frac{\alpha^2\sigma^2[1+(1-\alpha\mu)^D]}{{B_{\mathrm{low}}}[1-(1-\alpha\mu)^2]}\\
&+\frac{(1-\alpha\mu)^{2D}L^2}{[1-(1-\alpha\mu)^D]^2\mu^2T}
\sum_{k=0}^{T-2}\E\|X_{k+1}-X_k\|_\F^2.
\end{align*}
Finally,
\begin{align*}
&1-(1-\alpha\mu)^D\ge\alpha\mu>0,\\
&\frac{(1-\alpha\mu)^{2D}}{1-(1-\alpha\mu)^D}\le\frac1{\alpha\mu},\\
&\frac{(1-\alpha\mu)^{2D}L^2}{[1-(1-\alpha\mu)^D]^2\mu^2}\le\frac{L^2}{\alpha^2\mu^4},\\
&\frac{\alpha^2\sigma^2[1+(1-\alpha\mu)^D]}{{B_{\mathrm{low}}}[1-(1-\alpha\mu)^2]}
=\frac{\alpha\sigma^2[1+(1-\alpha\mu)^D]}{\mu {B_{\mathrm{low}}}(2-\alpha\mu)}
\le\frac{2\alpha\sigma^2}{\mu {B_{\mathrm{low}}}}.
\end{align*}
These bounds first give the movement-dependent estimate
\begin{equation}\label{eq:loweravg}
\frac1T\sum_{k=0}^{T-1}\E\|Y_k^D-Y^\star(X_k)\|_\F^2
\le{}\frac{\|Y_0-Y^\star(X_0)\|_\F^2}{\alpha\mu T}
+\frac{2\alpha\sigma^2}{\mu {B_{\mathrm{low}}}} +\frac{L^2}{\alpha^2\mu^4T}\sum_{k=0}^{T-2}\E\|X_{k+1}-X_k\|_\F^2.
\end{equation}
For exact polar, $\|O_k\|_\F^2=\operatorname{rank}(M_k)$; for finite NS, $\|O_k\|_\F^2\le\operatorname{rank}(M_k)$.
Hence $\|X_{k+1}-X_k\|_\F^2 =\beta^2\|O_k\|_\F^2 \le\beta^2\operatorname{rank}(M_k)\le\beta^2r$.
Substituting into~\eqref{eq:loweravg} and using $(T-1)/T\le1$ proves~\eqref{eq:lower-rank-uniform}.
We retain~\eqref{eq:loweravg} for the joint tracking argument.
Empty sums are zero.
For finite $T$, bounded UL increments for either spectral update,~\eqref{eq:Ylip}, finite initialization, and the conditional noise bound imply finite second moments by the displayed recursion.
\end{proof}

\begin{lemma}[Hypergradient-momentum tracking]\label{lem:momentum}
In the expressions below, $\E\widehat{\nabla\Phi}(X_k)$ averages only the fresh hypergradient batches with $X_k,Y_k^D$ fixed, as in Lemma~\ref{lem:hg}; an outer expectation also averages the iterates.
For $0<\eta<1/L$ and $0\le\theta<1$, $\|M_k\|_\F\le \left({C_f}+\frac{L{C_f}}{\mu}\right)$, and
\begin{align}\label{eq:momentumavg}
&\frac1T\sum_{k=0}^{T-1}\E\|M_k-\nabla\Phi(X_k)\|_\F^2
\le{}\frac{4\left({C_f}+\frac{L{C_f}}{\mu}\right)^2}{T(1-\theta)}
+\frac{6L_\Phi^2}{T(1-\theta)^2}\sum_{k=0}^{T-2}\E\|X_{k+1}-X_k\|_\F^2\nonumber\\
&+\frac6T\sum_{k=0}^{T-1}\E\|\E\widehat{\nabla\Phi}(X_k)-\nabla\Phi(X_k)\|_\F^2\nonumber\\
&+(1-\theta)\left({C_f}+\frac{L{C_f}}{\mu}\right)^2.
\end{align}
\end{lemma}
\begin{proof}
The sample derivative bounds and~\eqref{eq:vbound} imply
\begin{align}\label{eq:hgbound}
\|\widehat{\nabla\Phi}(X_k)\|_\F
&\le\|\nabla_XF(\mathcal D_\F)\|_\F
+\|\nabla_{XY}^2G(\mathcal D_G)\|_{\op}\|v_Q\|_\F
\le {C_f}+\frac{L{C_f}}{\mu}
\quad\text{almost surely}.
\end{align}

By $M_{-1}=0$,~\eqref{eq:hgbound}, and induction, $\|M_k\|_\F\le\theta\|M_{k-1}\|_\F+(1-\theta)\|\widehat{\nabla\Phi}(X_k)\|_\F \le\left({C_f}+\frac{L{C_f}}{\mu}\right)$.
For $k\ge1$, subtracting the current hypergradient from the momentum recursion gives
\begin{align*}
&M_k-\nabla\Phi(X_k)={}\theta[M_{k-1}-\nabla\Phi(X_{k-1})]
+\theta[\nabla\Phi(X_{k-1})-\nabla\Phi(X_k)]\\
&+(1-\theta)[\E\widehat{\nabla\Phi}(X_k)-\nabla\Phi(X_k)]\\
&+(1-\theta)[\widehat{\nabla\Phi}(X_k)-\E\widehat{\nabla\Phi}(X_k)].
\end{align*}
Condition on all iterates and samples generated before the fresh UL batches at iteration $k$.
The first three terms are then fixed.
By fresh sampling, the conditional mean of the estimator is $\E\widehat{\nabla\Phi}(X_k)$, so its centered noise has zero cross term with their sum.
Taking total expectations gives
\begin{align*}
&\E\|M_k-\nabla\Phi(X_k)\|_\F^2
={}\E\Big\|\theta[M_{k-1}-\nabla\Phi(X_{k-1})]
+\theta[\nabla\Phi(X_{k-1})-\nabla\Phi(X_k)]\\
&\hspace{4em}+(1-\theta)[\E\widehat{\nabla\Phi}(X_k)-\nabla\Phi(X_k)]\Big\|_\F^2\\
&+(1-\theta)^2\E\|\widehat{\nabla\Phi}(X_k)-\E\widehat{\nabla\Phi}(X_k)\|_\F^2.
\end{align*}
Young's inequality with weight $(1-\theta)/2$, followed by $\|U+V\|_\F^2\le2\|U\|_\F^2+2\|V\|_\F^2$, gives
\begin{align*}
&\Big\|\theta[M_{k-1}-\nabla\Phi(X_{k-1})]
+\theta[\nabla\Phi(X_{k-1})-\nabla\Phi(X_k)]\\
&\qquad+(1-\theta)[\E\widehat{\nabla\Phi}(X_k)-\nabla\Phi(X_k)]\Big\|_\F^2\\
&\le\left(1+\frac{1-\theta}{2}\right)\theta^2\|M_{k-1}-\nabla\Phi(X_{k-1})\|_\F^2\\
&\quad+2\left(1+\frac2{1-\theta}\right)\theta^2\|\nabla\Phi(X_{k-1})-\nabla\Phi(X_k)\|_\F^2\\
&\quad+2\left(1+\frac2{1-\theta}\right)(1-\theta)^2\|\E\widehat{\nabla\Phi}(X_k)-\nabla\Phi(X_k)\|_\F^2.
\end{align*}
For $0\le\theta<1$,
\begin{align*}
&\left(1+\frac{1-\theta}{2}\right)\theta^2\le\theta,\\
&2\left(1+\frac2{1-\theta}\right)\theta^2\le2+\frac4{1-\theta}\le\frac6{1-\theta},\\
&2\left(1+\frac2{1-\theta}\right)(1-\theta)^2=2(1-\theta)^2+4(1-\theta)\le6(1-\theta).
\end{align*}
For the next display, condition on $X_k,Y_k^D$ and take all expectations only over the fresh hypergradient batches.
The variance identity and~\eqref{eq:hgbound} give
\begin{equation}\label{eq:hgvar}
\E[\|\widehat{\nabla\Phi}(X_k)-\E\widehat{\nabla\Phi}(X_k)\|_\F^2] =\E[\|\widehat{\nabla\Phi}(X_k)\|_\F^2]
-\|\E\widehat{\nabla\Phi}(X_k)\|_\F^2
\le\left({C_f}+\frac{L{C_f}}{\mu}\right)^2.
\end{equation}
Using~\eqref{eq:gradLip} and taking total expectations in~\eqref{eq:hgvar},
\begin{align*}
&\E\|M_k-\nabla\Phi(X_k)\|_\F^2
\le{}\theta\E\|M_{k-1}-\nabla\Phi(X_{k-1})\|_\F^2
+\frac{6L_\Phi^2}{1-\theta}\E\|X_k-X_{k-1}\|_\F^2\\
&+6(1-\theta)\E\|\E\widehat{\nabla\Phi}(X_k)-\nabla\Phi(X_k)\|_\F^2\\
&+(1-\theta)^2\left({C_f}+\frac{L{C_f}}{\mu}\right)^2,\\
&\E\|M_0-\nabla\Phi(X_0)\|_\F^2\le4\left({C_f}+\frac{L{C_f}}{\mu}\right)^2.
\end{align*}
Summing for $1\le k<T$, adding the initial error, and extending nonnegative sums,
\begin{align*}
&\sum_{k=0}^{T-1}\E\|M_k-\nabla\Phi(X_k)\|_\F^2
\le{}4\left({C_f}+\frac{L{C_f}}{\mu}\right)^2+\theta\sum_{k=0}^{T-1}\E\|M_k-\nabla\Phi(X_k)\|_\F^2\\
&+\frac{6L_\Phi^2}{1-\theta}\sum_{k=0}^{T-2}\E\|X_{k+1}-X_k\|_\F^2\\
&+6(1-\theta)\sum_{k=0}^{T-1}\E\|\E\widehat{\nabla\Phi}(X_k)-\nabla\Phi(X_k)\|_\F^2\\
&+T(1-\theta)^2\left({C_f}+\frac{L{C_f}}{\mu}\right)^2.
\end{align*}
Rearranging and dividing by $T(1-\theta)>0$ gives~\eqref{eq:momentumavg}.
The argument also includes $\theta=0$.
\end{proof}

\begin{proof}[\normalfont\bfseries Proof of Lemma~\ref{lem:joint-tracking}]
\mbox{}\par\nobreak\noindent
For exact polar, the supported singular values of $O_k$ are all one.
For finite NS, they lie in $(0,1]$, with no new singular directions, as shown in the proof of Lemma~\ref{lem:NS}.
Both maps send zero to zero.
Thus $\|Y^\star(X_{k+1})-Y^\star(X_k)\|_\F^2 \le\frac{L^2\beta^2}{\mu^2}\|O_k\|_\F^2 \le\frac{L^2\beta^2}{\mu^2}\operatorname{rank}(M_k)$, where the first inequality uses the LL-solution Lipschitz bound~\eqref{eq:Ylip}.
Applying the warm-started LL tracking bound~\eqref{eq:loweravg} with $X_{k+1}-X_k=-\beta O_k$ gives~\eqref{eq:rank-lower}; retaining $\|O_k\|_\F^2$ gives the corresponding movement bound.

Substitute the conditional hypergradient bias bound~\eqref{eq:hgbias} into the momentum bound~\eqref{eq:momentumavg}.
The LL-error contribution is $12L_\Phi^2 T^{-1}\sum_{k=0}^{T-1}\E\|Y_k^D-Y^\star(X_k)\|_\F^2$.
Applying the movement-dependent bound in~\eqref{eq:loweravg} to this term gives the bound below.
The first movement coefficient comes from changes in the true hypergradient; the second comes from LL-solution tracking.
The rank identities above give the remaining claims.

The same argument also gives the following momentum bound used in the convergence proof:
\begin{align}\label{eq:joint-tracking}
&\frac1T\sum_{k=0}^{T-1}\E\|M_k-\nabla\Phi(X_k)\|_\F^2\nonumber\\
&\le \frac{4({C_f}+L{C_f}/\mu)^2}{T(1-\theta)}
+(1-\theta)({C_f}+L{C_f}/\mu)^2
+\frac{12L_\Phi^2\|Y_0-Y^\star(X_0)\|_\F^2}{\alpha\mu T}\nonumber\\
&\quad+\frac{24L_\Phi^2\alpha\sigma^2}{\mu {B_{\mathrm{low}}}}
+\frac{12L^2{C_f}^2}{\mu^2}(1-\eta\mu)^{2Q+2}\nonumber\\
&\quad+\frac{\beta^2}{T}
\left[\frac{6L_\Phi^2}{(1-\theta)^2}
+\frac{12L_\Phi^2L^2}{\alpha^2\mu^4}\right]
\sum_{k=0}^{T-2}\E\|O_k\|_\F^2.
\end{align}
For exact polar, $\|O_k\|_\F^2=\operatorname{rank}(M_k)$, so the last sum is exactly $\sum_{k=0}^{T-2}\E\operatorname{rank}(M_k)$.
For finite NS, $\|O_k\|_\F^2\le\operatorname{rank}(M_k)$, and replacing the last sum by the same rank sum remains valid.
Empty sums are zero.

\end{proof}

\begin{lemma}[Tracking under the Theorem~\ref{thm:main} stepsize schedule]\label{lem:spectral-tracking}
Suppose Assumptions~\ref{ass:convex}--\ref{ass:noise} hold.
Choose the parameters of Theorem~\ref{thm:main} and use either UL update.
Then
\begin{align}\label{eq:spectral-tracking}
&\frac1T\sum_{k=0}^{T-1}\E\|M_k-\nabla\Phi(X_k)\|_\F^2\nonumber\\
&\le T^{-1/2}\Bigg[
5\left({C_f}+\frac{L{C_f}}{\mu}\right)^2
+\frac{12L_\Phi^2L}{\mu}\|Y_0-Y^\star(X_0)\|_\F^2
+\frac{24L_\Phi^2\sigma^2}{L\mu {B_{\mathrm{low}}}}\nonumber\\
&\hspace{5em}+\frac{12L^2{C_f}^2}{\mu^2}
+r\beta_0^2L_\Phi^2\left(6+\frac{12L^4}{\mu^4}\right)\Bigg].
\end{align}
\end{lemma}
\begin{proof}
By orthonormality of the support singular vectors, $\|\polar(M_k)\|_{\op}\le1$, and $\|\polar(M_k)\|_\F^2=\operatorname{rank}(M_k)\le r$.
If $M_k=0$, both norms are zero.
Lemma~\ref{lem:NS} gives $\|\NS(M_k,q)\|_{\op}\le1$, hence $\|\NS(M_k,q)\|_\F^2\le r$.
Thus, for either update, $\|X_{k+1}-X_k\|_\F^2=\beta^2\|O_k\|_\F^2\le r\beta^2$, $\frac1T\sum_{k=0}^{T-2}\E\|X_{k+1}-X_k\|_\F^2\le\frac{T-1}{T}r\beta^2\le r\beta^2$.
The schedule satisfies $0<\alpha\le1/L$, $0<\eta<1/L$, $0\le\theta<1$, and $D\ge1$.
Thus~\eqref{eq:joint-tracking} applies to either update.
Bound its final sum by $(T-1)r\le Tr$ and substitute the schedule below.
All these expectations are finite by the LL recursion, bounded derivative estimates, and the bounded increments above.
Direct substitution yields
\begin{align*}
\frac1{T(1-\theta)}&=1-\theta=T^{-1/2},&
\frac1{\alpha\mu T}&=\frac L\mu T^{-1/2},\\
\alpha&=L^{-1}T^{-1/2},&
(1-\eta\mu)^{2Q+2}&\le(1-\eta\mu)^{2Q}\le T^{-1/2},\\
\frac{r\beta^2}{(1-\theta)^2}&=r\beta_0^2T^{-1/2},&
\frac{r\beta^2}{\alpha^2}&=r\beta_0^2L^2T^{-1/2}.
\end{align*}
Substituting these bounds gives~\eqref{eq:spectral-tracking}.
\end{proof}

\begin{lemma}[One-step progress of exact-polar \alg]\label{lem:svd-descent}
For the exact-polar update $X_{k+1}=X_k-\beta\polar(M_k)$ with $\beta>0$,
\begin{align}\label{eq:svd-descent}
\E\|\nabla\Phi(X_k)\|_*
\le{}&\frac{\E\Phi(X_k)-\E\Phi(X_{k+1})}{\beta}
+2\E\|M_k-\nabla\Phi(X_k)\|_*
+\frac{L_\Phi r\beta}{2}.
\end{align}
\end{lemma}
\begin{proof}
By the smoothness of $\Phi$ and $\|\polar(M_k)\|_\F^2\le r$,
\begin{align*}
\Phi(X_{k+1})
&\le\Phi(X_k)-\beta\langle\nabla\Phi(X_k),\polar(M_k)\rangle
+\frac{L_\Phi r\beta^2}{2}\\
&=\Phi(X_k)-\beta\|M_k\|_*
+\beta\langle M_k-\nabla\Phi(X_k),\polar(M_k)\rangle
+\frac{L_\Phi r\beta^2}{2}\\
&\le\Phi(X_k)-\beta\|\nabla\Phi(X_k)\|_*
+2\beta\|M_k-\nabla\Phi(X_k)\|_*
+\frac{L_\Phi r\beta^2}{2}.
\end{align*}
The equality uses $\langle M_k,\polar(M_k)\rangle=\|M_k\|_*$.
The last inequality follows from spectral--nuclear duality, $\|\polar(M_k)\|_{\op}\le1$, and the triangle inequality.
Rearranging and taking expectations proves~\eqref{eq:svd-descent}.
\end{proof}

\begin{proof}[\normalfont\bfseries Proof of Theorem~\ref{thm:main}: exact-polar case]
\mbox{}\par\nobreak\noindent
Averaging~\eqref{eq:svd-descent} over $k=0,\ldots,T-1$ and using $\E\Phi(X_T)\ge\Phi_{\inf}$ gives
\begin{align*}
\frac1T\sum_{k=0}^{T-1}\E\|\nabla\Phi(X_k)\|_*
&\le\frac{\Phi(X_0)-\Phi_{\inf}}{\beta T}
+\frac{2}{T}\sum_{k=0}^{T-1}\E\|M_k-\nabla\Phi(X_k)\|_*
+\frac{L_\Phi r\beta}{2}.
\end{align*}
For every matrix $A\in\mathbb R^{m\times n}$, Cauchy--Schwarz on its singular values gives $\|A\|_*\le\sqrt r\|A\|_\F$.
Applying Cauchy--Schwarz first in probability and then to the finite sum,
\begin{align*}
\frac1T\sum_{k=0}^{T-1}\E\|M_k-\nabla\Phi(X_k)\|_*
&\le\frac{\sqrt r}{T}\sum_{k=0}^{T-1}
\bigl(\E\|M_k-\nabla\Phi(X_k)\|_\F^2\bigr)^{1/2}\\
&\le\left[\frac rT\sum_{k=0}^{T-1}\E\|M_k-\nabla\Phi(X_k)\|_\F^2\right]^{1/2}.
\end{align*}
Using Lemma~\ref{lem:spectral-tracking}, $1/(\beta T)=\beta_0^{-1}T^{-1/4}$, and $T^{-3/4}\le T^{-1/4}$ proves the bound with
\begin{align}\label{eq:C-svd}
C_{\mathrm{pol}}={}&\frac{\Phi(X_0)-\Phi_{\inf}}{\beta_0}
+\frac{L_\Phi r\beta_0}{2}\nonumber\\
&+2\Bigg\{r\Bigg[
5\left({C_f}+\frac{L{C_f}}{\mu}\right)^2
+\frac{12L_\Phi^2L}{\mu}\|Y_0-Y^\star(X_0)\|_\F^2
+\frac{24L_\Phi^2\sigma^2}{L\mu {B_{\mathrm{low}}}}\nonumber\\
&\hspace{5em}+\frac{12L^2{C_f}^2}{\mu^2}
+r\beta_0^2L_\Phi^2\left(6+\frac{12L^4}{\mu^4}\right)\Bigg]\Bigg\}^{1/2}.
\end{align}
The constant is independent of $T$ and is valid for every fixed $\beta_0>0$.
Its three additive terms account for the initial UL objective gap, the smoothness cost of the UL updates, and the momentum tracking error bounded in Lemma~\ref{lem:spectral-tracking}, respectively.
The expectations of objective values exist, since $\|X_k-X_0\|_\F\le k\beta\sqrt r$, and $|\Phi(X_k)-\Phi(X_0)|\le({C_f}+L{C_f}/\mu)k\beta\sqrt r$.
Independence and uniformity of $R$ give
\[
\E\|\nabla\Phi(X_R)\|_*=
\sum_{k=0}^{T-1}\mathbb P(R=k)\E\|\nabla\Phi(X_k)\|_*
=\frac1T\sum_{k=0}^{T-1}\E\|\nabla\Phi(X_k)\|_*.
\]
This proves Theorem~\ref{thm:main}(i).
Choosing $T\ge\max\{2,\lceil(C_{\mathrm{pol}}/\epsilon)^4\rceil\}$ then ensures $\E\|\nabla\Phi(X_R)\|_*\le\epsilon$.
\end{proof}

\section{Convergence proof for finite-NS \alg}\label{app:ns}
We follow the polar-error descent argument in Appendix B of \citet{kim2026muon}, using the bilevel tracking bound established under the schedule of Theorem~\ref{thm:main}.
The Frobenius smoothness used here contributes the factor $r$ in the quadratic remainder.

The finite-NS map and its implementation are given in Appendix~\ref{app:algorithm-details}.

\begin{lemma}[One-step progress of finite-NS \alg]\label{lem:NS}\label{lem:epsilon-descent}
Suppose Assumptions~\ref{ass:convex}--\ref{ass:noise} hold.
For Algorithm~\ref{alg:bimuon} with $0<\eta<1/L$, $0\le\theta<1$, $\beta>0$, an integer $q\ge0$, and $O_k=\NS(M_k,q)$,
\begin{equation}\label{eq:NSnorm}
\|O_k\|_{\op}\le1,\qquad \|O_k\|_\F^2\le\operatorname{rank}(M_k)\le r.
\end{equation}
If~\eqref{eq:epsilon-def} holds, then
\begin{align}\label{eq:epsilon-descent}
(1-\varepsilon_q)\E\|\nabla\Phi(X_k)\|_*
\le{}&\frac{\E\Phi(X_k)-\E\Phi(X_{k+1})}{\beta}
+2\E\|M_k-\nabla\Phi(X_k)\|_*
+\frac{L_\Phi r\beta}{2}.
\end{align}
\end{lemma}
\begin{proof}
For the Taylor polynomial defined in Appendix~\ref{app:algorithm-details}, Proposition~1 of \citet{kim2026muon} establishes preservation of the unit spectral ball.
In scalar form, the properties needed here are $1\le p(x)\le p(0)\quad(0\le x\le1)$, and $p(x)\le x^{-1/2}\quad(0<x\le1)$.
The latter also follows directly by bounding the finite Taylor sum by the full series for $x^{-1/2}$.
Each positive initial singular value of $O^{[0]}$ is at most one; the NS recurrence sends $s$ to $sp(s^2)\in[s,1]$, while zero stays zero.
This proves~\eqref{eq:NSnorm}, including zero and rank-deficient momentum.
The NS map is continuous and hence measurable.

Insert the polar factor as in the single-level proof:
\begin{align*}
\langle M_k,O_k\rangle
&=\|M_k\|_*+\langle M_k,O_k-\polar(M_k)\rangle
\ge(1-\varepsilon_q)\|M_k\|_*,\\
\langle\nabla\Phi(X_k),O_k\rangle
&\ge(1-\varepsilon_q)\|M_k\|_*-\|M_k-\nabla\Phi(X_k)\|_*\\
&\ge(1-\varepsilon_q)\|\nabla\Phi(X_k)\|_*
-(2-\varepsilon_q)\|M_k-\nabla\Phi(X_k)\|_*\\
&\ge(1-\varepsilon_q)\|\nabla\Phi(X_k)\|_*
-2\|M_k-\nabla\Phi(X_k)\|_*.
\end{align*}
The first two inequalities use spectral--nuclear duality and the last two use the triangle inequality and $0\le\varepsilon_q<1$.
Applying~\eqref{eq:descent} gives $\Phi(X_{k+1})\le{}\Phi(X_k)-\beta(1-\varepsilon_q)\|\nabla\Phi(X_k)\|_* +2\beta\|M_k-\nabla\Phi(X_k)\|_* +\frac{L_\Phi r\beta^2}{2}$.
Rearranging and taking expectations yields~\eqref{eq:epsilon-descent}.
These expectations are finite: the updates have norm at most $\beta\sqrt r$, and both the true hypergradient and the momentum are bounded by $C_f+LC_f/\mu$.
\end{proof}

\paragraph{Polar approximation.}
Lemma~\ref{lem:ns-approximation} restates the Taylor NS residual and polar-error bounds of \citet[Theorem~2, Lemmas~1--3, and Corollary~5]{kim2026muon}.

Let $O_k^{[j]}$ denote the $j$th inner iterate of Algorithm~\ref{alg:ns} at outer iteration $k$, and let $\Pi_k$ be the orthogonal projector onto the column space of $M_k$.
Define $\delta_{k,j}:=\|\Pi_k-O_k^{[j]}(O_k^{[j]})^\top\|_{\op}$; when $M_k=0$, take $\Pi_k=0$, so this residual is zero.
For the degree-$d$ Taylor iteration, the residual satisfies $\delta_{k,q}\le\delta_{k,0}^{(d+1)^q}$.
Under the uniform initial bound in Lemma~\ref{lem:ns-approximation}, the resulting approximation factor satisfies $\frac{1}{1-\varepsilon_q}\le\frac{1}{\sqrt{1-\delta_0^{(d+1)^q}}}$ and approaches one doubly exponentially with $q$ for fixed $\delta_0<1$.
We apply these bounds on the positive singular support; Algorithm~\ref{alg:ns} places its initial singular values in $(0,1]$ and preserves that support.
Zero input is mapped to zero.
The cited result gives $\NS(M,q)\to\polar(M)$ for each fixed input $M$, but this pointwise conclusion does not imply the deterministic run-wise error condition~\eqref{eq:epsilon-def} used in Theorem~\ref{thm:main}.

\begin{proof}[\normalfont\bfseries Proof of Theorem~\ref{thm:main}: finite-NS case]
\mbox{}\par\nobreak\noindent
The schedule in Theorem~\ref{thm:main} satisfies the hypotheses of Lemma~\ref{lem:spectral-tracking}.
Averaging~\eqref{eq:epsilon-descent} and using $\E\Phi(X_T)\ge\Phi_{\inf}$ gives
\begin{align*}
&(1-\varepsilon_q)\frac1T\sum_{k=0}^{T-1}\E\|\nabla\Phi(X_k)\|_*\\
&\le\frac{\Phi(X_0)-\Phi_{\inf}}{\beta T}
+\frac2T\sum_{k=0}^{T-1}\E\|M_k-\nabla\Phi(X_k)\|_*
+\frac{L_\Phi r\beta}{2}\\
&\le\frac{\Phi(X_0)-\Phi_{\inf}}{\beta_0}T^{-1/4}
+2\left[\frac rT\sum_{k=0}^{T-1}\E\|M_k-\nabla\Phi(X_k)\|_\F^2\right]^{1/2}
+\frac{L_\Phi r\beta_0}{2}T^{-3/4}\\
&\le C_{\mathrm{pol}}T^{-1/4}.
\end{align*}
The second inequality uses $\|A\|_*\le\sqrt r\|A\|_\F$ and Cauchy--Schwarz in probability and over the finite sum.
The last inequality follows from~\eqref{eq:spectral-tracking} and~\eqref{eq:C-svd}.
Division by the positive deterministic $1-\varepsilon_q$ and independent uniform selection of $R$ prove Theorem~\ref{thm:main}(ii).
No unbiased-hypergradient assumption is introduced: LL-solution and Neumann bias are already included in Lemma~\ref{lem:spectral-tracking}.
\end{proof}

\paragraph{Uniform approximation condition.}
Theorem~\ref{thm:main}(ii) assumes a deterministic bound
$\varepsilon_q<1$ for the specified run.
A fixed NS depth alone does not guarantee that this bound remains
uniform as $T$ varies under the capped normalization in Algorithm~\ref{alg:ns}.
Thus, the horizon-uniform rate requires a common
$\varepsilon_q<1$ independent of $T$.
\paragraph{Computational complexity.}
An LL mini-batch mean uses ${B_{\mathrm{low}}}$ individual sample gradients per step.
Each outer iteration performs $D$ such steps.
Algorithm~\ref{alg:neumann} uses exactly $Q$ Hessian batches of size ${B_H}$, each applied to one recursively computed matrix; it includes $Q+1$ terms but does not require $Q+1$ Hessian batches.
The UL loss supplies two partial derivatives from a batch of size ${B_f}$.
The mixed action uses ${B_{xy}}$ samples.
For exact polar, or NS with a common polar-error bound below one, the derivative counts are as follows.
The orders in the final column hold with that error bound and the problem parameters fixed:
\begin{center}\begin{tabular}{lll}\toprule
Operation & Count through $T$ iterations & Order at nuclear accuracy $\epsilon$\\\midrule
Lower $\nabla_YG$ & $TD B_{\mathrm{low}}$ & $O(\epsilon^{-4})$\\
Upper partial $\nabla_XF$ & $T{B_f}$ & $O(\epsilon^{-4})$\\
Upper partial $\nabla_YF$ & $T{B_f}$ & $O(\epsilon^{-4})$\\
Hessian-vector actions & $TQ B_H$ & $O(\epsilon^{-4}\log(1/\epsilon))$\\
Mixed derivative actions & $T{B_{xy}}$ & $O(\epsilon^{-4})$\\\bottomrule
\end{tabular}\end{center}
The table counts derivative evaluations separately.
Finite NS additionally requires $q$ matrix-multiplication iterations per
UL update, while exact polar requires a matrix decomposition. These
matrix-arithmetic costs are separate from the stochastic derivative
complexity reported above.

\section{Proofs and Details for Blockwise Matrix \alg}\label{sec:blocks}\label{app:multiblock}
This appendix proves the blockwise LL sensitivity bound and the relative stepsize allocation rule in Section~\ref{sec:blockwise}, and provides additional details on the joint sensitivity condition and its finite-$D$ interpretation.

\begin{proposition}[Blockwise sensitivity of the LL solution]
\label{prop:block-sensitivity}
Suppose that $g(X,\cdot)$ is differentiable and $\mu$-strongly convex for every $X$, and that, for constants $L_{XY,b}>0$, the joint blockwise Lipschitz condition
\begin{equation}\label{eq:block-xy-lipschitz}
\|\nabla_Y g(X,Y)-\nabla_Y g(X',Y)\|_\F^2
\leq\sum_{b=1}^J L_{XY,b}^2\|X^{(b)}-X'^{(b)}\|_\F^2
\end{equation}
holds for all $X,X',Y$.
Then
\begin{equation}
\left\|
Y^\star(X)-Y^\star(X')
\right\|_\F^2
\le
\frac{1}{\mu^2}
\sum_{b=1}^{J}
L_{XY,b}^2
\left\|
X^{(b)}-X'^{(b)}
\right\|_\F^2.
\label{eq:block-y-star}
\end{equation}
\end{proposition}

\begin{proof}[\normalfont\bfseries Proof of Proposition~\ref{prop:block-sensitivity}]
\mbox{}\par\nobreak\noindent
Differentiability and strong convexity on the full LL Euclidean space give a unique minimizer: the strong-convexity quadratic lower bound is coercive, so continuity gives existence, and strong convexity gives uniqueness.
The two LL gradients vanish at their respective minimizers.
Strong monotonicity at fixed $X$ and Frobenius Cauchy--Schwarz give
\begin{align*}
\mu\|Y^\star(X)-Y^\star(X')\|_\F^2
&\le\langle\nabla_Yg(X,Y^\star(X'))-\nabla_Yg(X,Y^\star(X)),Y^\star(X')-Y^\star(X)\rangle\\
&\le\|\nabla_Yg(X,Y^\star(X'))-\nabla_Yg(X',Y^\star(X'))\|_\F
\|Y^\star(X)-Y^\star(X')\|_\F.
\end{align*}
If the minimizers coincide,~\eqref{eq:block-y-star} is immediate.
Otherwise divide by their positive distance, square, and apply~\eqref{eq:block-xy-lipschitz}.
Substitution of the update in~\eqref{eq:block-update} gives the intermediate bound $\|Y^\star(X_{k+1})-Y^\star(X_k)\|_\F^2 \le\frac{\beta^2}{\mu^2}\sum_{b=1}^J L_{XY,b}^2\omega_{b,k}^2\|O_k^{(b)}\|_\F^2$.
For exact polar, the squared norm equals the supported rank.
For the Taylor NS map, the singular values start in $[0,1]$, remain there under the specified polynomial, and no new supported direction appears; hence the squared norm is at most that rank, including zero momentum.
This is the same bound used in Lemma~\ref{lem:NS}.
Bounding each rank by $r_b$ and using $w_b=L_{XY,b}^2r_b/\mu^2$ yields the LL target-movement bound
\begin{equation}\label{eq:block-drift}
\|Y^\star(X_{k+1})-Y^\star(X_k)\|_\F^2
\leq\beta^2\sum_{b=1}^J w_b\omega_{b,k}^2.
\end{equation}
\end{proof}

At an outer transition, warm starting gives $Y_{k+1}^0=Y_k^D$.
For any $t>0$, Young's inequality gives
\[
\|Y_k^D-Y^\star(X_{k+1})\|_\F^2
\le(1+t)\|Y_k^D-Y^\star(X_k)\|_\F^2
+(1+t^{-1})\|Y^\star(X_{k+1})-Y^\star(X_k)\|_\F^2.
\]
Under the same LL contraction/noise conditions used in Lemma~\ref{lem:lower}, the finite-$D$ recursion contracts this new initial error in conditional expectation and adds its sampling term.
Equation~\eqref{eq:block-drift} controls the target-movement contribution; it does not imply monotonicity of the realized error because the two vector errors may reinforce or cancel.

\paragraph{Proof of the allocation rule in~\eqref{eq:block-allocation-solution}.}
For $a_{b,k}\ge0$ and fixed $w_b>0$, weighted Cauchy--Schwarz yields $\sum_b a_{b,k}\omega_{b,k} \le\sqrt{\sum_b a_{b,k}^2/w_b}\sqrt{\sum_b w_b\omega_{b,k}^2} \le\sqrt{\sum_b a_{b,k}^2/w_b}$.
When some score is positive, equality requires a saturated constraint and proportionality between $\sqrt{w_b}\omega_{b,k}$ and $a_{b,k}/\sqrt{w_b}$, giving~\eqref{eq:block-allocation-solution} and uniqueness.
All-zero scores make every feasible allocation optimal; zero is a convenient choice.
For exact polar, $\langle M,\polar(M)\rangle=\|M\|_*\ge0$.
Taylor NS has the same supported singular vectors and nonnegative singular values, so its score is also nonnegative.

With exactly the weights $w_b=L_{XY,b}^2r_b/\mu^2$, equation~\eqref{eq:block-drift} becomes $\|Y^\star(X_{k+1})-Y^\star(X_k)\|_\F^2\le\beta^2\sum_b w_b\omega_{b,k}^2\le\beta^2$ under the constraint in~\eqref{eq:block-drift-budget}.
Since $\beta>0$, taking square roots proves the stated drift bound.
Positivity of $L_{XY,b}$, $r_b$ and $\mu$ ensures $w_b>0$.

\section{Experimental Details and Supplementary Results}
\label{app:experiments}

This appendix provides implementation and evaluation details for the experiments in Section~\ref{sec:experiments}, {together with supplementary synthetic and controlled studies.}

\subsection{CIFAR-10 Representation Learning: Experimental Details}
\label{app:real_data}

\paragraph{Setup.}
The experiment in Section~\ref{sec:real_data} uses a fixed class-balanced pool of 1,500 CIFAR-10 \citep{krizhevsky2009learning} training images, sampled with pool seed 20270921.
A frozen ResNet-18 \citep{he2016deep} pretrained on ImageNet \citep{deng2009imagenet}, with its final classifier removed, provides 512-dimensional features using the preprocessing associated with torchvision's \texttt{IMAGENET1K\_V1} weights.
For each problem seed $s$, a stratified split assigns 100 images per class to LL training and 50 images per class to UL validation, giving 1,000 LL-training and 500 UL-validation samples.
The split generator uses seed $20000+s$.

Features are divided by a single RMS scalar computed from the LL training features, without whitening, centering, or coordinatewise rescaling.
The learned representation is $\tanh(zX^\top)$, $X\in\mathbb R^{256\times512}$, followed by a linear classifier $Y\in\mathbb R^{10\times256}$.
The LL objective is the mean training cross-entropy plus $0.01\|Y\|_\F^2/2$, and the UL objective is the mean validation cross-entropy.
We initialize $X_0$ as $\frac{0.1}{\sqrt{512}}$ times a standard Gaussian matrix generated using seed $30000+s$, and $Y_0=M_{-1}=0$.
All methods use the same split, initialization, and stochastic training stream for each problem seed.

\paragraph{Methods and hyperparameters.}
{stocBiO+Adam applies Adam~\citep{kingma2015adam} to the raw stochastic hypergradient with first- and second-moment decay rates $0.9$ and $0.999$, a denominator offset of $10^{-8}$, and bias correction.}
All methods use $D=5$ warm-started LL SGD steps with $\alpha=0.1$, $Q=4$ stochastic Hessian--vector products (five Neumann terms) with $\eta=0.2$, and fresh mini-batches of size 64; {all except stocBiO+Adam use EMA coefficient $\theta=0.9$.}
Each run performs 60 UL iterations in float64.

The scalar-normalized baselines in Section~\ref{sec:real_data} floor their normalization denominators at $10^{-30}$.
Exact-polar \alg applies the supported polar factor to the stochastic hypergradient momentum, retaining singular values larger than $10^{-14}\sigma_{\max}$.
NS-\alg uses the practical Frobenius-normalized cubic iteration $Z_0=\frac{M_k}{\|M_k\|_\F}$, $Z_{j+1}=\frac{3}{2}Z_j-\frac{1}{2}Z_jZ_j^\top Z_j$, $j=0,\ldots,11$, with zero output for zero momentum and the smaller Gram matrix used in the implementation.
Thus, the real-data experiment uses a practical Frobenius-normalized NS implementation rather than the capped normalization used in the theoretical analysis.

Every method uses the same stochastic derivative budget: 300 LL-gradient mini-batch evaluations, 60 UL partial-gradient pairs, 240 Hessian--vector products, and 60 mixed-derivative products.
Evaluation refits and spectral diagnostics are performed separately and do not affect the training state.

\paragraph{Tuning and evaluation.}
Stepsizes are selected using discovery seeds 0 and 1.
For each method, the initial candidates are calibrated to first-update Frobenius norms $\{0.025,\,0.1,\,0.4\}$, pooled over the two discovery seeds.
The candidate range is extended by factors of four when the selected value lies on an outer boundary.
The common candidate set also includes $\beta=0.14792005232672834$.
The selected stepsize minimizes the mean final optimized UL validation loss over the two discovery seeds, with ties resolved in favor of the smaller stepsize.
All selected stepsizes are frozen before evaluation on seeds 20--24, with 60 UL iterations per run.
{The selected UL stepsizes are $\beta=1.0101525$ for stocBiO+EMA, $\beta=0.14792005$ for OpNorm, FrobNorm, exact-polar \alg and NS-\alg, and $\beta=0.00110509$ for stocBiO+Adam.}

\paragraph{Optimized UL objective.}
At every logged $X_k$, including initialization and all 60 UL updates, the LL classifier is independently refit by L-BFGS \citep{liu1989limited}.
The refit uses at most 300 iterations, gradient tolerance $10^{-10}$, change tolerance $10^{-14}$, history size 50, and a strong-Wolfe line search.
The optimized UL validation loss reported in Section~\ref{sec:real_data} is the validation cross-entropy evaluated after this independent LL refit.

Figure~\ref{fig:realdata_ul}(a) reports means over evaluation seeds 20--24.
The inset compares exact-polar and NS-\alg over the final five iterations, with standard-error bands.
Table~\ref{tab:realdata_final} reports the final means and sample standard deviations.

\paragraph{Momentum-spectrum diagnostic.}
Figure~\ref{fig:realdata_momentum_spectrum}(b) uses the raw EMA momentum $M_k$ before the UL transformation.
{For stocBiO+Adam, $M_k$ is its uncorrected first-moment estimate, before bias correction and coordinatewise second-moment scaling.}
Let $\{\sigma_i\}$ denote its singular values.
The spectrum-imbalance statistic is $\frac{\mathrm{P90}(\{\sigma_i\})}{\max\{\mathrm{P10}(\{\sigma_i\}),10^{-12},10^{-6}\sigma_{\max}\}}$, where $\mathrm{P90}$ and $\mathrm{P10}$ denote the 90th and 10th percentiles, respectively.
A larger value indicates a more unequal singular spectrum.
For each method and iteration, Figure~\ref{fig:realdata_momentum_spectrum}(b) reports the median across the five evaluation seeds, with first- and third-quartile bands.
The diagnostic is computed on the input momentum $M_k$, rather than on the transformed UL direction.

\begin{table}[!htbp]
\centering
\small

\begin{tabular}{lr}
\toprule
UL update & Optimized UL validation loss\\
\midrule
stocBiO+EMA & $.182175\pm.019159$\\
OpNorm & $.199307\pm.127671$\\
FrobNorm & $.189266\pm.021947$\\
{stocBiO+Adam} & {$.174908\pm.018090$}\\
Exact-polar \alg & $.033333\pm.005441$\\
NS-\alg & $.037494\pm.007741$\\
\bottomrule
\end{tabular}
\caption{Final optimized UL validation loss after 60 updates on CIFAR-10.
Uncertainties are sample standard deviations over seeds 20--24.}
\label{tab:realdata_final}
\end{table}

{\paragraph{Comparison with stocBiO+Adam.}
Table~\ref{tab:realdata_final} shows that stocBiO+Adam is the strongest baseline; both \alg variants achieve a lower final UL validation loss on all five paired evaluation seeds.}

\paragraph{Runtime protocol.}
Figure~\ref{fig:gpu_timing}(c) reports a separate benchmark on an NVIDIA A100-SXM4-40GB using PyTorch 1.12.1+cu116, CUDA 11.6, float64, with TF32 disabled.
The normalized cubic NS depth is fixed to 12 and exact-polar and NS-\alg use the common UL stepsize $0.14792005232672834$.

Single-transformation timing uses 12 momentum snapshots collected at iterations 1, 30, and 60 from seeds 20 and 21.
Both transformations receive the same input matrices, with five warmup calls followed by seven timed rounds of ten calls per method in randomized method order.

The 60-update timing uses seeds 20--24 with three repetitions per seed and includes the LL solver, stochastic hypergradient estimator, EMA update, batching, and UL transformation.
Feature extraction, LL evaluation refits, spectral diagnostics, and logging are excluded.
Reported speedups are medians of paired exact-polar/NS timing ratios.
The marginal median runtimes are 16.91 ms versus 1.19 ms per transformation and 1.385 s versus 0.411 s over 60 updates for exact polar and NS-12, respectively.

\subsection{CIFAR-100 Blockwise Allocation: Experimental Details}
\label{app:cifar100_blockwise}

\paragraph{Setup.}
The experiment in Section~\ref{sec:cifar100_blockwise} uses all 50,000 CIFAR-100 \citep{krizhevsky2009learning} training images.
A fixed class-stratified split assigns 400 images per class to LL training and 100 images per class to UL validation, giving 40,000/10,000 samples, with split seed 20270922.

A frozen ResNet-50 \citep{he2016deep} pretrained on ImageNet \citep{deng2009imagenet} with torchvision \texttt{IMAGENET1K\_V2} weights provides 2,048-dimensional features.
Preprocessing uses bilinear resizing to 232 pixels, a 224-pixel center crop, and channel normalization with means $(0.485,0.456,0.406)$ and standard deviations $(0.229,0.224,0.225)$.
All features are divided by the LL-training RMS scalar $0.3188708678543272$, without whitening, centering, or coordinatewise rescaling.
All methods use the same cached features and fixed data split.

\paragraph{Controlled coupling and exact block sensitivity.}
Write $X,Y\in\mathbb R^{100\times2048}$ as four column blocks $X^{(b)},Y^{(b)}\in\mathbb R^{100\times512}$, $b=1,\ldots,4$.
Set $c_b=\varrho^{(b-1)/3}$, $C=\operatorname{blkdiag}(c_1I_{512},\ldots,c_4I_{512})$.
For frozen features $z_i$ and labels $t_i$, define $\ell_{\mathcal S}(Y)=\frac{1}{|\mathcal S|}\sum_{i\in\mathcal S}\operatorname{CE}(Yz_i,t_i)$.
The bilevel objectives are $g(X,Y)=\ell_{\mathrm{tr}}(Y)+\frac{\mu}{2}\|Y-XC\|_\F^2$, $f(X,Y)=\ell_{\mathrm{val}}(Y)$, $\mu=0.01$.
The shared softmax cross-entropy couples all classifier blocks.
Moreover, $\nabla_{XY}^2g(X,Y)[V]=-\mu VC$, and $\|\nabla_Y g(X,Y)-\nabla_Y g(X',Y)\|_\F^2=\mu^2\sum_{b=1}^4 c_b^2\|X^{(b)}-X'^{(b)}\|_\F^2$.
Hence $L_{XY,b}=\mu c_b$, $r_b=100$, $w_b=100c_b^2$.
The sensitivity ratio is $\varrho=\frac{\max_b c_b}{\min_b c_b}$, and the corresponding largest-to-smallest ratio among the $w_b$ is $\varrho^2$.

\paragraph{Methods and hyperparameters.}
All runs start from $X_0=Y_0=M_{-1}=0$ and use a seed-dependent mini-batch stream initialized with seed $100000+s$.
We use $D=5$ warm-started LL updates, $Q=4$ Hessian--vector products (five Neumann terms), batch size 64, and EMA coefficient $\theta=0.9$ without bias correction.

The LL feature bound $L=\frac{1}{2}\max_{i\in\mathrm{tr}}\|z_i\|_2^2+\mu$ determines $\alpha=\min\{0.1,1/L\}$, $\eta=\min\{0.2,1/L\}$, which both equal $0.0004690390185025686$ in this experiment.
These parameters are shared across all allocation rules and all values of $\varrho$.

Each block uses the exact supported polar direction $O_k^{(b)}=\polar(M_k^{(b)})$, with singular values below $10^{-14}\sigma_{\max}$ removed.
Let $\widetilde\omega_{b,k}$ denote the raw allocation weight.
Every rule is normalized by $\omega_{b,k}=\frac{\widetilde\omega_{b,k}}{\sqrt{\sum_c w_c\widetilde\omega_{c,k}^2}}$, followed by $X_{k+1}^{(b)}=X_k^{(b)}-\beta\omega_{b,k}O_k^{(b)}$.
Positive raw weights are first divided by their maximum for numerical stability, which does not change the normalized allocation.
If all progress scores vanish, the update is set to zero; otherwise $\sum_b w_b\omega_{b,k}^2=1$.

Each 60-iteration run uses 300 LL-gradient mini-batch evaluations, 240 Hessian--vector products, 60 mixed-derivative products, and 60 UL partial-gradient pairs.
Optimization uses float64 with TF32 disabled on an NVIDIA A100-SXM4-40GB with PyTorch 2.11.0+cu128, torchvision 0.26.0+cu128, and CUDA 12.8.
Frozen-backbone feature extraction is performed separately in float32.

\paragraph{Tuning and evaluation.}
A single common UL stepsize is selected from $\beta\in\{0.001,0.01,0.1,1\}$ using 25-iteration discovery runs on seeds 31001 and 31002, all four allocation rules, and $\varrho\in\{1,10\}$.
The selected stepsize minimizes the mean final optimized UL validation loss over the discovery runs, with ties resolved in favor of the smaller value.
This gives $\beta=0.1$, which is frozen for all final evaluations.
Evaluation uses seeds 41001--41005, $\varrho\in\{1,3,10\}$, and 60 UL iterations.

\paragraph{Optimized UL objective.}
At initialization and every five UL iterations, a deterministic full-data L-BFGS solve approximates $Y^\star(X)$, warm-started from the previous evaluation reference.
The solve uses at most 300 iterations, learning rate 1, history size 30, strong-Wolfe line search, gradient tolerance $10^{-10}$, change tolerance $10^{-14}$, and full-objective accumulation in chunks of 10,000 images.
A reference solution is accepted when $\|\nabla_Yg(X,\widehat Y)\|_\F\le 10^{-5}$.
The reported optimized UL loss is $\ell_{\mathrm{val}}(\widehat Y)$.

Figure~\ref{fig:cifar100_blockwise}(a) reports means and sample-standard-deviation bands over the five evaluation seeds at $\varrho=10$.
Figure~\ref{fig:cifar100_blockwise}(b) reports the final mean optimized UL loss at $\varrho\in\{1,3,10\}$ with sample-standard-deviation error bars.

\begin{table}[!htbp]
\centering\small

\begin{tabular}{crrrr}
\toprule
$\varrho$ & Uniform & Progress-only & Sensitivity-only & Proposed\\
\midrule
1 & $.822789\pm.002729$ & $.822794\pm.002733$ & $.822789\pm.002729$ & $.822794\pm.002733$\\
3 & $.829264\pm.002747$ & $.841830\pm.002435$ & $.830485\pm.002349$ & $.822794\pm.002733$\\
10 & $.843067\pm.002478$ & $.860136\pm.001990$ & $.844844\pm.001927$ & $.822794\pm.002733$\\
\bottomrule
\end{tabular}
\caption{Final optimized UL validation loss for CIFAR-100 blockwise allocation with common $\beta=0.1$.
Entries are means and sample standard deviations over five evaluation seeds.}
\label{tab:cifar100_blockwise}
\end{table}

For both $\varrho=3$ and $\varrho=10$, the proposed allocation attains a lower final optimized UL loss than the three alternative allocation rules on all five evaluation seeds.

\paragraph{Scale invariance of the proposed rule.}
The nearly identical proposed results across different values of $\varrho$ follow from the scale structure of this controlled construction.
Writing $Z^{(b)}=c_bX^{(b)}$, $M_k^{(b)}=c_bN_k^{(b)}$, and using $w_b=100c_b^2$, the factor $c_b$ cancels from the effective proposed update in $Z^{(b)}$.
Consequently, under the shared zero initialization and random stream, the proposed trajectories in $(Z,Y)$ are invariant to $\varrho$.

\subsection{Supplementary Synthetic Representation Learning}
\label{app:nonlinear-confirmation}
\label{sec:nonlinear_representation}
\label{sec:exp_nonlinear}

\paragraph{Setup.}
We additionally study a controlled four-class nonlinear representation-learning problem with $X\in\mathbb R^{32\times16}$, $Y\in\mathbb R^{4\times32}$.
The representation is $\tanh(zX^\top)$.
The LL objective is training cross-entropy plus $0.01\|Y\|_\F^2/2$, and the UL objective is cross-entropy on a separate validation set.
The LL-training, UL-validation, and test sample sizes are 256, 128, and 128, respectively.

The teacher representation has rank 8.
Inputs are generated as $z=\xi U\operatorname{diag}(s)U^\top$, where $\xi$ is standard Gaussian, $U$ is a fixed orthogonal matrix, and the entries of $s$ are geometrically spaced with ratio $\kappa_{\mathrm{in}}$ and normalized such that $\sum_i s_i^2=16$.
Thus, the population input-covariance condition number is $\kappa_{\mathrm{in}}^2$ while the total input variance is fixed.
Labels are generated by the same low-rank tanh teacher construction for each value of $\kappa_{\mathrm{in}}$.

\begin{figure}[!htbp]
\setlength{\parskip}{0pt}
\setlength{\abovecaptionskip}{3pt}
\centering
\begin{minipage}[t]{.49\linewidth}
\centering
\includegraphics[width=\linewidth]{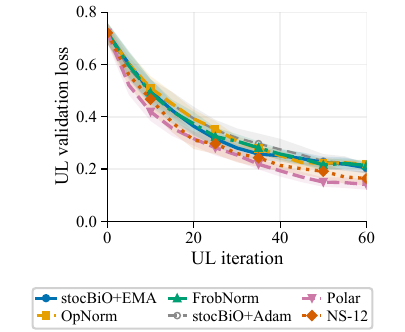}
\par\small(a) Selected task
\end{minipage}\hfill
\begin{minipage}[t]{.49\linewidth}
\centering
\includegraphics[width=\linewidth]{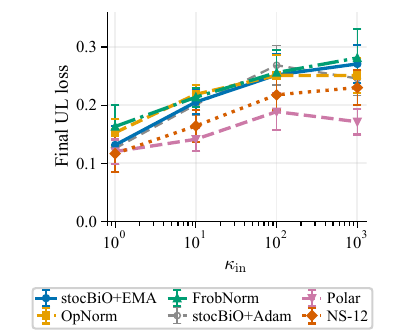}
\par\small(b) Input-anisotropy sweep
\end{minipage}
\caption{Supplementary nonlinear representation-learning results{, including stocBiO+Adam}.
Left: the frozen configuration at $\kappa_{\mathrm{in}}=10$.
Right: the input-anisotropy sweep, with method-specific UL stepsizes selected on discovery seeds at each value.
Bands and error bars show standard errors over five evaluation seeds.
The optimized UL validation loss is computed after independently refitting the LL classifier.}
\label{fig:nonlinear-supplement}
\end{figure}

\paragraph{Methods and hyperparameters.}
{We compare the same six methods as in Section~\ref{sec:real_data}, with stocBiO+Adam configured as in Appendix~\ref{app:real_data}.
All six methods use the same warm-started stochastic bilevel estimator with $D=5$, $Q=4$, $\alpha=0.1$, $\eta=0.2$, and $B=64$; all except stocBiO+Adam use $\theta=0.9$.}
NS-\alg uses the cubic Taylor iteration with degree $d=1$, depth $q=12$, and initialization $\frac{M}{\max\{1,\|M\|_\F\}}$.
OpNorm and FrobNorm use denominator floor $10^{-30}$.
Exact-polar \alg applies the supported polar factor to the same stochastic hypergradient momentum.

\paragraph{Tuning and evaluation.}
Hyperparameters are selected using discovery seeds 0 and 1.
The selected representative setting uses $\kappa_{\mathrm{in}}=10$.
All methods use the same 17-point UL-stepsize candidate grid, with a separate stepsize selected for each method.
The grid is
\[
\begin{aligned}
\beta\in\{&0.00625,\ 0.0125,\ 0.025,\ 0.05,\ 0.1,\ 0.2,\ 0.4,\ 0.8,\ 1.6,\\
&3.2,\ 6.4,\ 12.8,\ 25.6,\ 51.2,\ 102.4,\ 204.8,\ 409.6\}.
\end{aligned}
\]
The resulting solver parameters and stepsizes are frozen before evaluation on seeds 10--14.
{In the input-anisotropy sweep below, the stepsizes are selected in the same way for each $\kappa_{\mathrm{in}}\in\{1,10,100,1000\}$; for stocBiO+Adam, this gives $\beta=0.2$, $0.2$, $0.2$, and $0.1$, respectively.}
All evaluation runs use 60 UL iterations without retuning or early stopping.

At each logged $X$, the LL classifier is independently refit using L-BFGS, and the primary metric is the resulting optimized UL validation cross-entropy.
Curves report means and standard errors over the five evaluation seeds.

\paragraph{Results.}
At $\kappa_{\mathrm{in}}=10$, exact-polar \alg achieves final optimized UL validation loss $0.1414\pm0.0452$, NS-\alg achieves $0.1643\pm0.0602$, and stocBiO+EMA achieves $0.2054\pm0.0469$, where uncertainties are sample standard deviations over the five evaluation seeds.
The two \alg variants therefore provide an additional controlled comparison of spectral reshaping against the same scalar-normalized baselines used in the real-data experiment.

We also vary the input anisotropy while keeping the representation rank, solver depths, NS normalization, and stopping rule fixed.
Figure~\ref{fig:nonlinear-supplement} reports the selected $\kappa_{\mathrm{in}}=10$ configuration together with the anisotropy sweep{, and Table~\ref{tab:synthetic_final} lists the final losses.
Exact-polar \alg reaches a lower final optimized UL validation loss than stocBiO+Adam on all five seeds for every $\kappa_{\mathrm{in}}\ge10$, and at $\kappa_{\mathrm{in}}=1$ the two \alg variants still have the lowest mean losses.}

\begin{table}[!htbp]
\centering\small
\setlength{\tabcolsep}{4pt}

\begin{tabular}{lrrrr}
\toprule
Method & $\kappa_{\mathrm{in}}=1$ & $10$ & $100$ & $1000$\\
\midrule
stocBiO+EMA & $0.131\pm0.074$ & $0.205\pm0.047$ & $0.252\pm0.080$ & $0.271\pm0.074$\\
OpNorm & $0.153\pm0.051$ & $0.219\pm0.037$ & $0.251\pm0.079$ & $0.251\pm0.067$\\
FrobNorm & $0.163\pm0.082$ & $0.214\pm0.037$ & $0.256\pm0.086$ & $0.281\pm0.112$\\
stocBiO+Adam & $0.126\pm0.063$ & $0.201\pm0.041$ & $0.269\pm0.077$ & $0.246\pm0.066$\\
Exact-polar \alg & $0.120\pm0.046$ & $\mathbf{0.141\pm0.045}$ & $\mathbf{0.189\pm0.071}$ & $\mathbf{0.171\pm0.049}$\\
NS-\alg & $\mathbf{0.117\pm0.072}$ & $0.164\pm0.060$ & $0.218\pm0.060$ & $0.230\pm0.067$\\
\bottomrule
\end{tabular}
\caption{Final optimized UL validation loss in the input-anisotropy sweep. Means and sample standard deviations over five evaluation seeds.}
\label{tab:synthetic_final}
\end{table}

\subsection{Supplementary Controlled Blockwise Allocation}
\label{app:blockwise_exp}
\label{sec:blockwise_exp}

\paragraph{Setup.}
We further study the blockwise allocation rule on a controlled quadratic bilevel problem with $J=4$ UL blocks $X^{(b)}\in\mathbb R^{16\times16}$ and an LL matrix $Y\in\mathbb R^{64\times16}$.
Let $P_b\in\mathbb R^{64\times16}$ embed block $b$ into its corresponding 16 rows, so that $P_b^\top P_c=0\, (b\neq c)$, $P_b^\top P_b=I$.
We use $g(X,Y)=\frac12\langle Y,HY\rangle-\left\langle\sum_{b=1}^4 c_bP_bX^{(b)},Y\right\rangle$, and $f(X,Y)=\frac12\|CY-R\|_\F^2$.

The LL curvature matrix is $H=U\operatorname{diag}(h)U^\top$, where $U$ is obtained by QR factorization of a standard Gaussian matrix and the 64 eigenvalues are logarithmically spaced from 1 to 10.
Thus $\mu=1$, $\kappa(H)=10$.
The dense matrix $C\in\mathbb R^{64\times64}$ has independent Gaussian-QR left and right orthogonal factors and singular values linearly spaced from 0.5 to 1.5.
We set $R=CY_{\mathrm{target}}$, where the entries of $Y_{\mathrm{target}}$ are independently drawn from $\mathcal N(0,1/64)$.

The exact LL solution is $Y^\star(X)=H^{-1}\sum_{b=1}^4 c_bP_bX^{(b)}$.
We set $c_b=\varrho^{(b-1)/3}$, so that $\varrho$ is the block-sensitivity ratio.
The orthogonal embeddings give $\|\nabla_Yg(X,Y)-\nabla_Yg(X',Y)\|_\F^2=\sum_{b=1}^4 c_b^2\|X^{(b)}-X'^{(b)}\|_\F^2$.
Hence $L_{XY,b}=c_b$, $r_b=16$, $w_b=16c_b^2$.
For each seed, $H$, $C$, and $R$ are fixed across all allocation rules and all values of $\varrho$.
All runs start from $X_0=0$, $Y_0=0$, $M_{-1}=0$.

\paragraph{Methods and hyperparameters.}
We use the deterministic full-gradient specialization of the same bilevel recurrence, with $D=5$, $\alpha=0.1$, $Q=4$, $\eta=0.1$, $\theta=0.9$.
For each block, $O_k^{(b)}=\polar(M_k^{(b)})$, $a_{b,k}=\langle M_k^{(b)},O_k^{(b)}\rangle$.
The supported polar implementation retains singular values larger than $10^{-14}$ times the largest singular value and maps zero momentum to zero.
All runs use 100 UL iterations and CPU float64 arithmetic.

The raw allocation weights are
\[
\widetilde\omega_{b,k}=\begin{cases}
1,&\text{Uniform},\\
a_{b,k},&\text{Progress-only},\\
1/w_b,&\text{Sensitivity-only},\\
a_{b,k}/w_b,&\text{Proposed}.
\end{cases}
\]
Every rule is normalized as $\omega_{b,k}=\frac{\widetilde\omega_{b,k}}{\sqrt{\sum_cw_c\widetilde\omega_{c,k}^2}}$, followed by $X_{k+1}^{(b)}=X_k^{(b)}-\beta\omega_{b,k}O_k^{(b)}$.
Positive raw weights are divided by their maximum before normalization for numerical stability.
If all progress scores vanish, the update is set to zero; otherwise $\sum_b w_b\omega_{b,k}^2=1$.

\begin{figure}[!htbp]
\centering
\includegraphics[width=\linewidth]{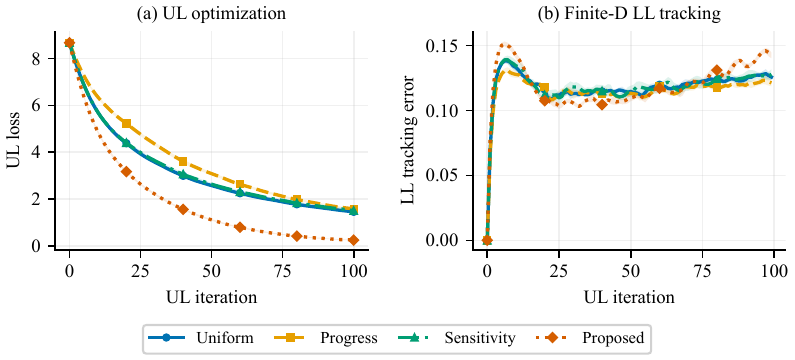}
\caption{Controlled blockwise relative stepsize allocation at $\varrho=10$.
All methods use the same global stepsize and theoretical LL-drift budget; only the relative allocation across matrix blocks differs. {Hence the LL tracking errors in (b) remain comparable, while the proposed rule attains the lowest UL loss in (a).}
Curves show means and standard-error bands over five evaluation seeds.}
\label{fig:blockwise_alloc}
\end{figure}

\paragraph{Tuning and evaluation.}
The pilot uses seeds 0 and 1, $\varrho\in\{1,10,30\}$, $\beta\in\{0.03,0.1,0.3,1.0\}$, for all four allocation rules.
One common $\beta$ is selected by minimizing the mean final-to-initial UL objective ratio over the pilot runs, with ties resolved in favor of the smaller stepsize.
This selects $\beta=0.3$.
The value is frozen for evaluation on seeds 100--104 and $\varrho\in\{1,3,10,30\}$.

The optimized UL objective is evaluated using the exact LL solution: $\Phi(X_k)=f(X_k,Y^\star(X_k))$.
We additionally report $\|Y_k^D-Y^\star(X_k)\|_\F$ as the finite-$D$ LL tracking error.
Curves and error bars report means and standard errors over the five paired evaluation seeds; table entries report sample standard deviations.

\paragraph{Results.}
Figure~\ref{fig:blockwise_alloc} reports the representative $\varrho=10$ experiment under the common global stepsize $\beta=0.3$.
The proposed allocation achieves lower UL objective values than Uniform, Progress-only, and Sensitivity-only under the same blockwise drift budget.

\begin{table}[!htbp]
\centering\small

\begin{tabular}{crrrr}
\toprule
$\varrho$ & Uniform & Progress-only & Sensitivity-only & Proposed\\
\midrule
1 & $.2536\pm.0258$ & $.2512\pm.0212$ & $.2536\pm.0258$ & $.2512\pm.0212$\\
3 & $.5217\pm.0408$ & $.4753\pm.0405$ & $.5750\pm.0868$ & $.2512\pm.0212$\\
10 & $1.4566\pm.0711$ & $1.5870\pm.0719$ & $1.5110\pm.1253$ & $.2512\pm.0212$\\
30 & $2.4444\pm.1239$ & $2.8098\pm.1317$ & $2.5259\pm.1470$ & $.2512\pm.0212$\\
\bottomrule
\end{tabular}
\caption{Final optimized UL objective $\Phi(X_{100})$ under the common $\beta=0.3$.
Entries are means and sample standard deviations over five evaluation seeds.}
\label{tab:blockwise_sweep}
\end{table}

\begin{figure}[!htbp]
\centering
\includegraphics[width=\linewidth]{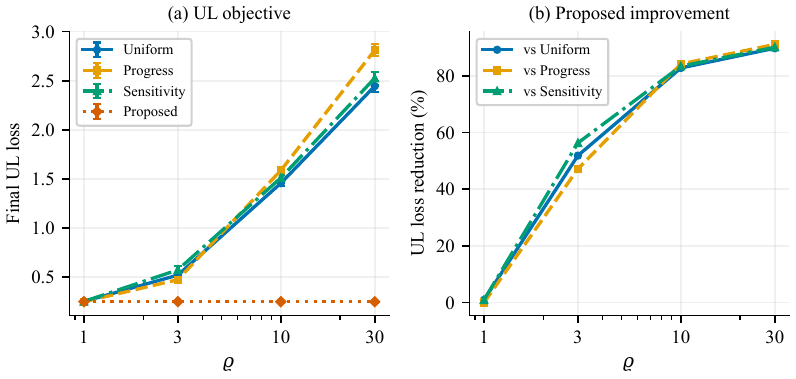}
\caption{Sensitivity-ratio sweep for controlled blockwise allocation with the common $\beta=0.3$.
Means and standard errors are computed over five evaluation seeds.
The same problem matrices and initialization are reused across $\varrho$ within each seed.}
\label{fig:blockwise_sweep}
\end{figure}

Table~\ref{tab:blockwise_sweep} and Figure~\ref{fig:blockwise_sweep} show the complete sensitivity-ratio sweep.
For $\varrho\in\{3,10,30\}$, the proposed allocation attains a lower final UL objective than all three alternatives on all five paired evaluation seeds.
At $\varrho=1$, Uniform and Sensitivity-only coincide, while Progress-only and Proposed coincide.

\paragraph{Scale invariance.}
The constant proposed trajectory across $\varrho$ follows from the scaling
structure of this construction. Writing
$Z^{(b)}=c_bX^{(b)}$ and $M_k^{(b)}=c_bN_k^{(b)}$ gives
$a_{b,k}=c_b\widetilde a_{b,k}$ and $w_b=16c_b^2$.
Positive scalar multiplication does not change the polar direction, and the
factor $c_b$ cancels from the effective proposed update in $Z^{(b)}$.
Therefore, under the shared zero initialization, the proposed $(Z,Y)$
trajectories are invariant to $\varrho$.

\FloatBarrier
{\subsection{CIFAR-100 Transformer Representation Learning: Experimental Details}
\label{app:transformer}

\paragraph{Setup.}
The experiment in Section~\ref{sec:transformer_main} uses the ImageNet-pretrained DeiT-S/16 Transformer~\citep{touvron2021deit,krizhevsky2009learning,deng2009imagenet} (official \texttt{deit\_small\_patch16\_224} checkpoint).
The UL variables are the fused QKV, attention-output and two MLP projection matrices of the last four Transformer blocks, with shapes $1152\times384$, $384\times384$, $1536\times384$ and $384\times1536$.
The 16 UL matrices contain 7,077,888 parameters.
The class-stratified LL-training and UL-validation splits contain 40,000 and 10,000 images, respectively, and each run uses 1,000 UL iterations.
The first eight blocks, all biases and LayerNorm parameters, and the tokens are frozen; their outputs are cached, and the features are recomputed through the trainable last four blocks at every UL iteration.
The LL variable is a linear classifier $Y\in\mathbb R^{100\times384}$ on the CLS features, trained with cross-entropy plus $0.01\|Y\|_\F^2/2$, so the LL problem is $0.01$-strongly convex; the UL objective is the validation cross-entropy.
Images are resized to 248 pixels and center-cropped to $224\times224$, without augmentation.
Each matrix receives its own UL update with a common stepsize.

\paragraph{Methods and tuning.}
We use the six UL updates defined in Section~\ref{sec:real_data}, with 12 cubic Newton--Schulz steps for NS-12.
All methods use the same stochastic implicit hypergradient estimator, with $D=5$ LL steps ($\alpha=0.1$), $Q=1024$ Hessian--vector products ($\eta=0.2$) and mini-batch size 64, and all except stocBiO+Adam use the EMA coefficient $\theta=0.9$.
stocBiO+Adam applies Adam~\citep{kingma2015adam} with first- and second-moment decay rates $0.9$ and $0.999$ and bias correction to the raw hypergradient.
For every method, the stepsize is selected from the same grid of relative first-update scales $\{10^{-5},3\times10^{-5},\ldots,3\times10^{-3}\}$, with at most two boundary extensions, by the mean final UL validation loss on two discovery seeds, and is then frozen for five fresh paired evaluation seeds.
The selected UL stepsizes are $\beta=0.00696$, $0.00577$, $0.0106$, $0.000159$, $0.00192$ and $0.00281$ for stocBiO+EMA, OpNorm, FrobNorm, stocBiO+Adam, Polar and NS-12, respectively.
The UL validation loss is evaluated every 250 iterations after refitting the LL classifier.
All runs use FP32 on one NVIDIA H100 NVL GPU.
Curves and shading in Figure~\ref{fig:transformer_progress} show means and sample standard deviations over these five seeds; wall-clock time is measured from the start of each run.
Time-to-threshold speedups in Section~\ref{sec:transformer_main} are means of paired timing ratios across the evaluation seeds.

\begin{table}[!htbp]
\centering\small
\setlength{\tabcolsep}{4pt}

\begin{tabular}{lrrrr}
\toprule
Method & Loop time (s) & Time / EMA & UL update (ms) & Memory (GiB)\\
\midrule
stocBiO+EMA & $4.421\pm0.020$ & 1.000 & $1.9\pm0.1$ & 16.02\\
OpNorm & $4.659\pm0.026$ & 1.054 & $251.6\pm1.5$ & 16.02\\
FrobNorm & $4.424\pm0.022$ & 1.001 & $2.3\pm0.1$ & 16.02\\
stocBiO+Adam & $4.427\pm0.025$ & 1.002 & $2.6\pm0.1$ & 16.05\\
Polar & $4.651\pm0.024$ & 1.052 & $253.4\pm1.5$ & 16.02\\
NS-12 & $4.431\pm0.022$ & 1.002 & $12.3\pm0.1$ & 16.02\\
\bottomrule
\end{tabular}
\caption{Transformer optimization cost: loop time and UL-update time per iteration (mean and sample SD), mean paired loop-time ratio to stocBiO+EMA, and peak allocated GPU memory.}
\label{tab:transformer_cost}
\end{table}

\paragraph{Runtime and memory.}
Table~\ref{tab:transformer_cost} reports the optimization loop, which comprises feature refresh, LL tracking, hypergradient estimation and the UL update, and excludes setup and evaluation.
The shared hypergradient computation dominates the loop time, so both \alg variants add little overhead: relative to stocBiO+EMA, the loop time is $1.052\times$ for Polar and $1.002\times$ for NS-12.
Both \alg variants use the same peak GPU memory as stocBiO+EMA (Table~\ref{tab:transformer_cost}), including a shared 14.09-GiB cache of frozen-prefix features; each stores one momentum matrix per block.

\FloatBarrier
\subsection{Controlled UL Condition Numbers and Spectrum Shapes}
\label{app:controlled_ul_shapes}

\paragraph{Setup.}
We use the quadratic bilevel model of Appendix~\ref{app:controlled-spectral} with $X,Y\in\mathbb R^{16\times16}$, $\mu=0.1$, zero targets, isotropic validation curvature and $\gamma=0$.
Then $\nabla\Phi(X)=\mu^2H_{\mathrm{LL}}^{-2}X$, and the UL Hessian $\mathcal H_{\mathrm{UL}}[\Delta]=A\Delta$ with $A=\mu^2H_{\mathrm{LL}}^{-2}$ satisfies $\kappa(\mathcal H_{\mathrm{UL}})=\kappa(H_{\mathrm{LL}})^2$.
We prescribe $\kappa(H_{\mathrm{LL}})\in\{1,10,100,300,1000\}$, i.e., $\kappa_{\mathrm{UL}}:=\kappa(\mathcal H_{\mathrm{UL}})\in\{1,10^2,10^4,9\times10^4,10^6\}$, and fix the largest eigenvalue of $A$ at $\lambda_{\max}=0.01$.
This range covers our CIFAR-10 task: at the final iterates of stocBiO+EMA and Exact-Polar \alg, the LL Hessian has $\kappa(H_{\mathrm{LL}})$ between 313 and 644 across the five evaluation seeds, which corresponds to $\kappa_{\mathrm{UL}}\approx10^5$--$4\times10^5$ in this model.
To test the effect of the spectrum shape at a fixed condition number, we use four spectra of $A$, with $\lambda_{\min}=\lambda_{\max}/\kappa_{\mathrm{UL}}$: (1) Bulk + outlier: one eigenvalue $\lambda_{\max}$ and 15 geometrically spaced in $[\lambda_{\min},\min\{10\lambda_{\min},\lambda_{\max}\}]$; (2) Low atom + outlier: one at $\lambda_{\max}$ and 15 at $\lambda_{\min}$; (3) Geometric: 16 geometrically spaced in $[\lambda_{\min},\lambda_{\max}]$; and (4) High atom + trough: 15 at $\lambda_{\max}$ and one at $\lambda_{\min}$.
The eigenbasis is a random orthogonal matrix, and $X_0$ is a Gaussian matrix with $\|X_0\|_\F=4$ that is not aligned with the eigenbasis; both are shared by all methods within each seed.

\paragraph{Methods and tuning.}
We compare hypergradient descent with EMA momentum (HG+EMA), FrobNorm, Exact-Polar \alg (Polar) and NS-12 \alg, all with the exact hypergradient, EMA coefficient 0.9, a constant stepsize and 1,000 UL iterations in float64.
For every method and regime, the stepsize is selected by the same rule on discovery seeds 3000 and 3001 (a coarse grid of first-update lengths followed by a common 17-point refinement), minimizing the mean relative gap $\Phi(X_k)/\Phi(X_0)$ over iterations 901--1000, and is frozen before evaluation on five new paired seeds.

\begin{table}[H]
\centering\small
\setlength{\tabcolsep}{3pt}

\begin{tabular}{crrrr}
\toprule
$\kappa_{\mathrm{UL}}$ & HG+EMA & FrobNorm & Polar & NS-12\\
\midrule
$1$ & $(4.86\pm0.00)10^{-17}$ & $(2.25\pm0.00)10^{-6}$ & $(6.88\pm0.40)10^{-6}$ & $(4.67\pm0.78)10^{-6}$\\
\midrule
\multicolumn{5}{l}{(1) Bulk + outlier}\\
$10^2$ & $(2.10\pm0.45)10^{-16}$ & $(2.20\pm0.66)10^{-5}$ & $(1.41\pm1.65)10^{-5}$ & $(6.51\pm1.65)10^{-6}$\\
$10^4$ & $(3.34\pm0.85)10^{-7}$ & $(2.33\pm0.53)10^{-5}$ & $(5.01\pm2.95)10^{-6}$ & $(6.36\pm1.56)10^{-6}$\\
$9\times10^4$ & $(6.66\pm1.73)10^{-5}$ & $(1.74\pm0.96)10^{-4}$ & $(7.78\pm8.82)10^{-6}$ & $(8.16\pm1.70)10^{-6}$\\
$10^6$ & $(5.52\pm1.49)10^{-5}$ & $(7.74\pm1.87)10^{-5}$ & $(8.49\pm2.69)10^{-6}$ & $(2.73\pm1.12)10^{-5}$\\
\midrule
\multicolumn{5}{l}{(2) Low atom + outlier}\\
$10^2$ & $(9.24\pm2.01)10^{-19}$ & $(2.24\pm1.13)10^{-5}$ & $(8.88\pm6.41)10^{-6}$ & $(3.53\pm1.41)10^{-6}$\\
$10^4$ & $(3.45\pm0.89)10^{-6}$ & $(2.64\pm0.51)10^{-5}$ & $(3.92\pm0.83)10^{-6}$ & $(5.99\pm0.56)10^{-6}$\\
$9\times10^4$ & $(1.08\pm0.28)10^{-4}$ & $(2.01\pm0.57)10^{-4}$ & $(1.63\pm0.37)10^{-5}$ & $(1.80\pm1.29)10^{-5}$\\
$10^6$ & $(1.82\pm0.47)10^{-5}$ & $(2.24\pm0.49)10^{-5}$ & $(4.07\pm1.28)10^{-6}$ & $(1.17\pm0.31)10^{-5}$\\
\midrule
\multicolumn{5}{l}{(3) Geometric}\\
$10^2$ & $(4.82\pm0.66)10^{-16}$ & $(6.02\pm0.54)10^{-6}$ & $(9.17\pm0.74)10^{-6}$ & $(8.78\pm5.28)10^{-6}$\\
$10^4$ & $(9.00\pm1.16)10^{-8}$ & $(2.22\pm0.63)10^{-5}$ & $(1.09\pm0.44)10^{-5}$ & $(1.20\pm0.13)10^{-5}$\\
$9\times10^4$ & $(7.33\pm0.47)10^{-6}$ & $(3.49\pm1.02)10^{-5}$ & $(1.17\pm0.19)10^{-5}$ & $(1.50\pm0.07)10^{-5}$\\
$10^6$ & $(1.03\pm0.08)10^{-5}$ & $(3.83\pm0.82)10^{-5}$ & $(1.72\pm0.40)10^{-5}$ & $(1.58\pm0.17)10^{-5}$\\
\midrule
\multicolumn{5}{l}{(4) High atom + trough}\\
$10^2$ & $(1.69\pm0.24)10^{-16}$ & $(3.92\pm3.03)10^{-6}$ & $(9.52\pm1.21)10^{-6}$ & $(6.17\pm0.87)10^{-6}$\\
$10^4$ & $(6.15\pm0.88)10^{-9}$ & $(1.55\pm1.28)10^{-5}$ & $(5.65\pm0.21)10^{-6}$ & $(7.94\pm1.08)10^{-6}$\\
$9\times10^4$ & $(3.42\pm0.49)10^{-7}$ & $(2.92\pm2.40)10^{-6}$ & $(2.72\pm3.31)10^{-5}$ & $(4.53\pm0.71)10^{-6}$\\
$10^6$ & $(6.10\pm0.87)10^{-8}$ & $(1.71\pm0.86)10^{-6}$ & $(2.71\pm3.30)10^{-5}$ & $(3.84\pm0.75)10^{-6}$\\
\bottomrule
\end{tabular}
\caption{Controlled UL spectrum study: relative UL objective gap $\Phi(X_k)/\Phi(X_0)$ averaged over iterations 901--1000, with mean and sample SD over five fresh paired seeds (a displayed zero SD is due to rounding). At $\kappa_{\mathrm{UL}}=1$, the four spectra coincide.}
\label{tab:shape_sweep}
\end{table}

\paragraph{Results.}
Table~\ref{tab:shape_sweep} shows that spectral updates help most for spectra (1) and (2), where a few large eigenvalues sit above a bulk of small ones.
For spectrum (1), increasing $\kappa(H_{\mathrm{LL}})$ from 1 to 300 worsens the HG+EMA gap by more than twelve orders of magnitude, while the Polar gap remains of order $10^{-5}$.
At $\kappa_{\mathrm{UL}}=9\times10^4$, both \alg variants outperform HG+EMA and FrobNorm on all five evaluation seeds, with NS-12 reducing the mean gap by factors of 8.2 and 21.3, respectively; at $\kappa_{\mathrm{UL}}=10^6$, both variants still have lower mean gaps than both baselines.
FrobNorm has a larger gap than HG+EMA throughout spectrum (1).
For $\kappa_{\mathrm{UL}}\le10^4$, and for spectra (3) and (4) at every $\kappa_{\mathrm{UL}}$, HG+EMA performs better: constant-step normalized updates stall at gaps of about $10^{-6}$ to $10^{-5}$.
}

\end{document}